\documentclass[lettersize,journal]{IEEEtran}

\usepackage{multirow}
\usepackage{booktabs}
\usepackage{graphicx}
\usepackage[normalem]{ulem}
\useunder{\uline}{\ul}{}

\usepackage{amsmath,amsfonts}
\usepackage{algorithm}
\usepackage{array}
\usepackage[caption=false,font=footnotesize,labelfont=rm,textfont=rm,captionskip=-6pt]{subfig}
\usepackage{textcomp}
\usepackage{stfloats}
\usepackage{url}
\usepackage{verbatim}
\usepackage{graphicx}
\usepackage{cite}
\usepackage{amssymb}
\usepackage{xcolor}
\usepackage{pifont}
\usepackage{algpseudocode}
\usepackage{color}
\usepackage{hyperref}
\hypersetup{
    colorlinks=true,
    linkcolor=red,
    filecolor=magenta,      
    urlcolor=blue,
    pdftitle={Overleaf Example},
    pdfpagemode=FullScreen,
    }

\begin{document}

\definecolor{blue}{RGB}{68,114,196}
\definecolor{orange}{RGB}{237,125,49}
\def\etal{\textit{et al.}}
\def\etc{\textit{etc}}
\def\ie{\textit{i.e.}}
\def\eg{\textit{e.g.}}

\title{{G\textsuperscript{2}Face: High-Fidelity Reversible Face Anonymization via Generative and Geometric Priors}

\author{Haoxin~Yang,
        Xuemiao~Xu,
        Cheng~Xu,
        Huaidong~Zhang,
        Jing Qin,
        Yi~Wang,\\
        Pheng-Ann Heng,~\IEEEmembership{Senior Member,~IEEE},
        Shengfeng He,~\IEEEmembership{Senior Member,~IEEE}
        }

\thanks{Manuscript received at 15 January 2024; revised at 8 April 2024 and 12 June 2024; accepted at 15 August 2024. This work is supported by China National Key R\&D Program (Grant No. 2023YFE0202700), Key-Area Research and Development Program of Guangzhou City (No. 2023B01J0022); Guangdong International Technology Cooperation Project (No. 2022A0505050009); the Guangdong Natural Science Funds for Distinguished Young Scholar under Grant 2023B1515020097; National Research Foundation Singapore under the AI Singapore Programme under Grant AISG3-GV-2023-011; Guangdong Basic and Applied Basic Research Foundation (No. 2023B1515120058); Collaborative research with world-leading research groups in The Hong Kong Polytechnic University (NO. P0052927). \textit{(Corresponding authors: Xuemiao Xu; Cheng Xu.)}}
\thanks{Haoxin~Yang and Xuemiao~Xu are with the School of Computer Science and Engineering, South China University of Technology, Guangzhou, China.  Xuemiao Xu is also with Guangdong Engineering Center for Large Model and GenAI Technology, State Key Laboratory of Subtropical Building and Urban Science. E-mail: yhx1996@outlook.com; xuemx@scut.edu.cn.}
\thanks{Cheng~Xu and Jing~Qin are with the Centre for Smart Health, The Hong Kong Polytechnic University, Hong Kong. E-mail: cschengxu@gmail.com; harry.qin@polyu.edu.hk.}
\thanks{Huaidong~Zhang is with the School of Future Technology, South China University of Technology, Guangzhou, China. E-mail: huaidongz@scut.edu.cn.}
\thanks{Yi~Wang is with the School of Computer Science and Network Security, Dongguan University of Technology, Dongguan, China. E-mail: wangyi@dgut.edu.cn.}
\thanks{Pheng-Ann Heng is with the Department of Computer Science and Engineering, The Chinese University of Hong Kong, Hong Kong, SAR, China. E-mail: pheng@cse.cuhk.edu.hk.}
\thanks{Shengfeng He is with the School of Computing and Information Systems, Singapore Management University, Singapore. E-mail: shengfenghe@smu.edu.sg.}}

\markboth{IEEE TRANSACTIONS ON INFORMATION FORENSICS AND SECURITY}%
{Yang \MakeLowercase{\textit{et al.}}:G\textsuperscript{2}Face: High-Fidelity Reversible Face Anonymization via Generative and Geometric Priors}


\maketitle

\begin{abstract}
Reversible face anonymization, unlike traditional face pixelization, seeks to replace sensitive identity information in facial images with synthesized alternatives, preserving privacy without sacrificing image clarity. Traditional methods, such as encoder-decoder networks, often result in significant loss of facial details due to their limited learning capacity. Additionally, relying on latent manipulation in pre-trained GANs can lead to changes in ID-irrelevant attributes, adversely affecting data utility due to GAN inversion inaccuracies.
This paper introduces G\textsuperscript{2}Face, which leverages both generative and geometric priors to enhance identity manipulation, achieving high-quality reversible face anonymization without compromising data utility. We utilize a 3D face model to extract geometric information from the input face, integrating it with a pre-trained GAN-based decoder. This synergy of generative and geometric priors allows the decoder to produce realistic anonymized faces with consistent geometry.
Moreover, multi-scale facial features are extracted from the original face and combined with the decoder using our novel identity-aware feature fusion blocks (IFF). This integration enables precise blending of the generated facial patterns with the original ID-irrelevant features, resulting in accurate identity manipulation.
Extensive experiments demonstrate that our method outperforms existing state-of-the-art techniques in face anonymization and recovery, while preserving high data utility. Code is available at \href{https://github.com/Harxis/G2Face}{\textit{https://github.com/Harxis/G2Face}}.

\begin{IEEEkeywords} 
Reversible face anonymization, generative prior, geometric prior, identity-aware feature fusion.
\end{IEEEkeywords}

\end{abstract} 
\section{Introduction}

Recent advancements in deep learning and computer vision have significantly enhanced everyday convenience, but they have also raised substantial security concerns. This is particularly evident in the widespread sharing of personal facial images on social media, which creates a considerable risk of privacy breaches in case of unauthorized access. To address this, a surge of research~\cite{hukkelaas2019deepprivacy,maximov2020ciagan,gafni2019live,gu2020password,barattin2023attribute,li2023riddle,wen2023divide} has focused on face anonymization technology, which aims to alter the identity in facial images while preserving other ID-irrelevant attributes~\cite{na2022mfim}, such as hairstyles, expressions, and background. This technology ensures the privacy of facial images, maintaining their usefulness for various downstream tasks like face detection, tracking, and landmark detection.

\begin{figure}[t!]
   \centering
   \includegraphics[width=0.95\linewidth]{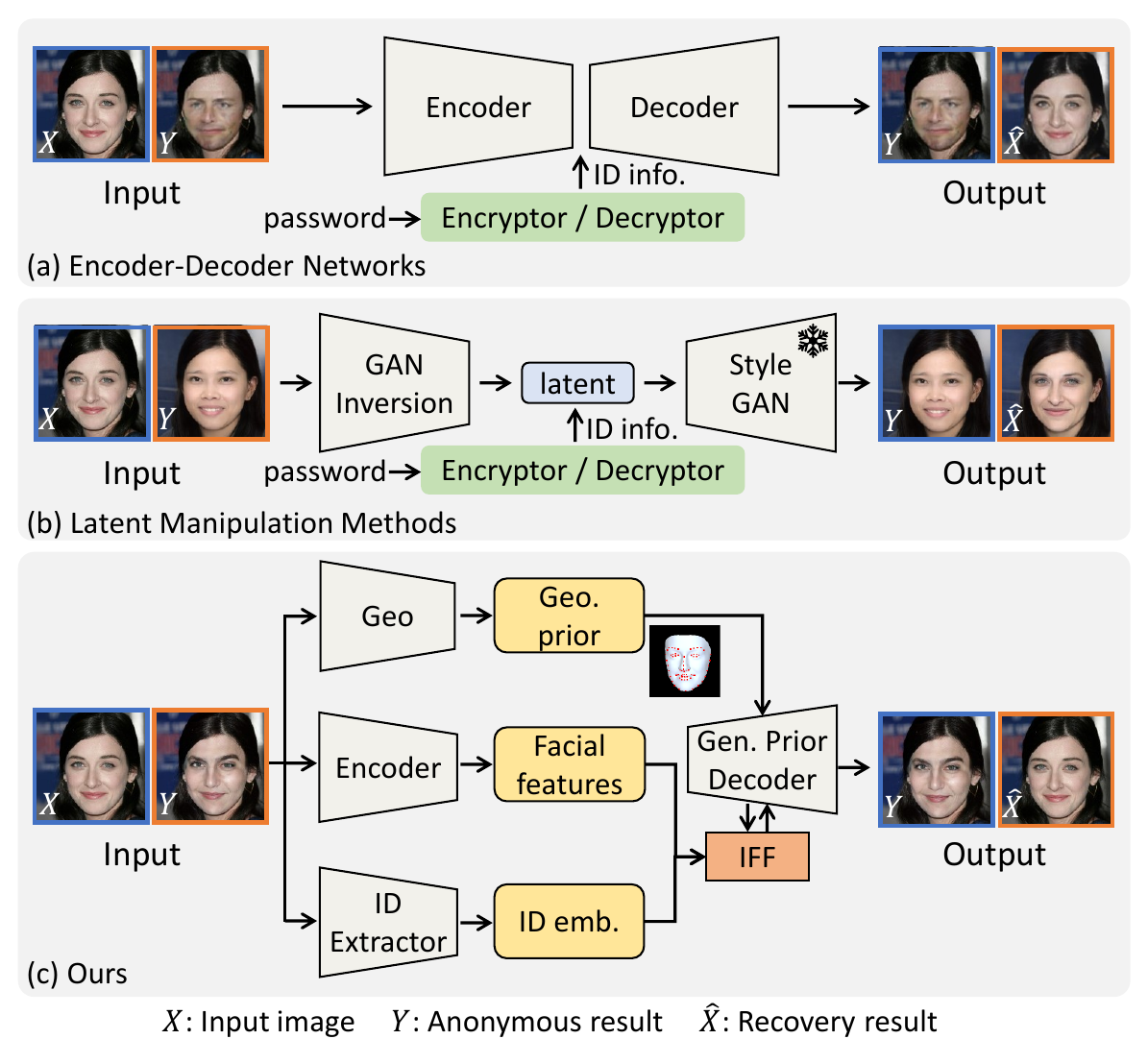}
   \caption{Current methods for face anonymization either employ a basic encoder-decoder model for identity modification (a) or modify the latent code using GAN inversion (b), both approaches often resulting in significant loss of facial detail or undesirable changes in ID-irrelevant attributes. Our proposed method intricately combines generative and geometric priors, enhanced by identity guidance, to refine the process of identity modification. This approach enables not only realistic face anonymization but also faithful recovery, all while effectively preserving the utility of the data (c).}
   \label{fig:teaser}
\end{figure}

Traditional face anonymization techniques, such as blurring and mosaicing, focus on pixel-level perturbations of facial regions. While effective in obscuring identity, these methods often compromise the clarity of images due to content distortion. 
The advent of deep learning, particularly generative adversarial networks (GANs)~\cite{goodfellow2020generative}, has led to the development of GAN-based face anonymization approaches that offer a better balance between identity concealment and image clarity \cite{maximov2020ciagan,gafni2019live,gu2020password,barattin2023attribute,li2023riddle, cao2021personalized,proencca2021uu,hukkelaas2019deepprivacy,chen2021perceptual,wen2023divide}. In practical scenarios, the capacity to reconstruct the original identity from anonymized faces is crucial, especially when individuals agree to share their data only with specific, authorized parties. Addressing this need, several studies\cite{gu2020password, proencca2021uu, cao2021personalized,li2023riddle,yuan2023invertible} have explored reversible face anonymization techniques. These techniques facilitate the generation of images using both anonymous and real IDs, enabling both the effective removal of identity information from the source image and the precise restoration of the original face from its anonymized version.

Despite significant advances, current face anonymization methods still face challenges in high-fidelity reversible face anonymization without compromising data utility. Traditional approaches primarily use an encoder-decoder architecture to learn a basic identity manipulation mapping~\cite{maximov2020ciagan, gu2020password, proencca2021uu, cao2021personalized,zhong2022delving}, which often leads to limited generative capacity and over-smoothed results with loss of facial details (refer to Fig.~\ref{fig:teaser}(a)). While some recent efforts have focused on enhancing the fidelity of face anonymization by manipulating the latent space of a pre-trained GAN model~\cite{li2023riddle, barattin2023attribute}, these methods struggle to preserve original ID-irrelevant attributes due to inaccuracies and limited control in the GAN-inversion process from input face to latent code (see Fig.~\ref{fig:teaser}(b)). Furthermore, many existing techniques do not adequately consider facial geometric constraints, leading to distortions in facial geometry and failure to maintain authentic facial shapes and expressions. This lack of geometric consideration significantly reduces the utility of anonymized faces for downstream tasks like landmark detection and expression recognition~\cite{maximov2020ciagan, gu2020password, li2023riddle, barattin2023attribute}.

To address the above challenges, we introduce G\textsuperscript{2}Face, a novel framework that seamlessly integrates generative priors and geometric priors to achieve realistic reversible face anonymization with faithful face shapes, while simultaneously preserving the ID-irrelevant attributes. Our framework is composed of three key modules: the \textit{Geometry-aware Identity Extraction (GIE)} module, the \textit{Dual Prior-guided Identity Modification (DPIM)} module, and the \textit{Password Extraction (PE)} module. In the anonymization process, the GIE module initially selects a random identity embedding and merges it with 3D geometry parameters derived from the input face, creating a geometry-informed identity embedding for target anonymization. The DPIM then utilizes this embedding to generate ID-relevant features, which are intricately blended with the original image's ID-irrelevant features through our \textit{identity-aware feature fusion block (IFF)}. This blending occurs on multiple scales, enhancing the anonymity and authenticity of the generated face images.
The final step involves the PE module, which extracts the original face's identity information embedded in the anonymized image. This extracted identity is then used to reconstruct the original face by inputting the embedding and anonymized image back into the DPIM. By strategically leveraging the strengths of both generative and geometric priors with precise identity guidance, G\textsuperscript{2}Face facilitates not only vivid anonymization but also accurate recovery, maintaining realistic facial details and shapes without altering ID-irrelevant attributes (Fig.~\ref{fig:teaser}(c)). Our extensive experiments validate our method's superiority over existing state-of-the-art techniques in terms of both privacy protection and data utility.

To summarize, our contributions are threefold:
\begin{enumerate}
\item We introduce G\textsuperscript{2}Face, a novel high-fidelity reversible face anonymization framework. It distinctively merges identity-aware generative with geometric priors for identity manipulation. This novel approach facilitates both face anonymization and recovery, delivering images with realistic facial details and shapes, and ensures the precise preservation of ID-irrelevant attributes.
\item We develop a specialized identity-aware feature fusion block. This block is designed to accurately integrate both generative and geometric priors into the primary facial features, guided by the identity information.
\item Our method not only exhibits superior performance in safeguarding identity privacy but also surpasses current state-of-the-art methods in enhancing data utility.
\end{enumerate}
\section{Related Work}
\subsection{Face Anonymization and Reversible Recovery}
Face anonymization seeks to protect the identity of a face image from being recognized by human observers or computer vision systems, by replacing the original identity with a randomly generated identity.
Recent studies have predominantly leveraged Generative Adversarial Networks~\cite{goodfellow2020generative} (GANs) for face anonymization. 
Specifically, DeepPrivacy~\cite{hukkelaas2019deepprivacy} obscures the face area and subsequently reconstructs the removed portion using image inpainting techniques, eliminating identity information from the image. CIAGAN~\cite{maximov2020ciagan} masks the face using predicted landmarks and subsequently inpaints the masked area, guided by an alternate identity.
FALCO~\cite{barattin2023attribute} drectly optimizes the latent codes of StyleGAN~\cite{karras2019style}, potentially altering identity-irrelevant attributes due to latent space coupling.
IDeudemon~\cite{wen2023divide} proposes an approach employing a ``divide and conquer" strategy to safeguard identity and utility incrementally while ensuring strong explainability.
PRO-Face~\cite{yuan2022pro} and PRO-Face C~\cite{yuan2024pro} propose to obscure the visual features of facial images while maintaining machine recognition capability. However, their methods decrease the utility of the images.

Additionally, there are also a few efforts to explore reversible anonymization solutions for achieving authenticated access to the original face.
FIT~\cite{gu2020password}, Pan et al.~\cite{pan2021multi} and Cao et al.~\cite{cao2021personalized}, RAPP~\cite{zhang2023rapp} proposed to obtain a new identity vector by encrypting the original identity vector with a password and obtain an anonymized image through a generation network, and then decrypt the original identity vector and restore the original image by entering the correct password.
PRO-Face S~\cite{yuan2023invertible} suggests their flow model can generate a face similar to a pre-obfuscated face by inputting both the original and pre-obfuscated faces, allowing for reversible generation.  Their main focus is on recovering anonymized faces, whereas our method enables simultaneous anonymization and recovery.
RiDDLE~\cite{li2023riddle} proposes a framework of reversible and diversified de-identification with a latent encryptor to generate an anonymized image and recover the original image by a pre-trained StyleGAN2 model~\cite{karras2020analyzing}.
Although face identity can be altered to some extent, the above methods either suffer from significant facial detail loss due to the inherent limited generation capacity of a simple encoder-decoder architecture or failure to preserve ID-irrelevant attributes caused by inaccurate inversion in pre-trained GAN models. In contrast, our method delicately combines the generative and geometric priors for identity manipulation, leading to realistic face anonymization and recovery with rich facial details and faithful facial shapes without impairing the ID-irrelevant facial attributes.

\subsection{Facial Prior}
\textbf{Generative Prior.}
In recent years, models from the Style-GAN series~\cite{karras2019style, karras2020analyzing, karras2021alias} have demonstrated exceptional capabilities in generating high-quality images, which give birth to a large number of image editing techniques by leveraging generative priors. StyleGAN~\cite{karras2019style} enables the generation of diverse images by controlling their style, StyleGAN2~\cite{karras2020analyzing} eliminates the AdaIN operation to effectively diminish artifacts, while StyleGAN3~\cite{karras2021alias} further enhances generation details by mapping discrete features into continuous ones. Despite the limited improvements in generation quality from StyleGAN3, considering the trade-off between computational/model complexities and generation performance, we have chosen StyleGAN2 as our facial prior module.
These methods typically utilize GAN inversion for image processing. First, the input image is encoded into the latent space of the StyleGAN to derive a latent code. This code is then manipulated for various image editing applications, such as face swapping~\cite{xu2022high}, face anonymization~\cite{li2023riddle, barattin2023attribute}, super-resolution~\cite{menon2020pulse, zhong2022faithful}, \etc.
However, a notable drawback of these methods is their infeasibility in finding the precise latent code of the input image, which causes inevitable changes in ID-irrelevant attributes. To tackle this issue, we present an identity-aware feature fusion block that fuses prior features and primitive features with the guidance of identity information. This innovative approach not only enhances the authenticity of the generated faces but also preserves facial attributes that are unrelated to identity, contributing to a more faithful and nuanced anonymization and recovery.

\textbf{3D geometric Prior.}
Apart from generative priors, 3D geometric facial priors are also widely used for enhancing 3D geometry awareness in face generation. In particular, 3D Morphable Models (3DMM), as a typical representation of the 3D shape and texture of a face image, has demonstrated its efficacy as 3D prior~\cite{blanz1999morphable}. For instance, it has proven successful in various face editing tasks such as face swapping~\cite{wang2021hififace}, expression and posture transfer~\cite{thies2016face2face, nirkin2018face}, and face super-resolution~\cite{hu2020face, hu2021face,liu20223dfp}.
These methods leverage the 3DMM coefficients of identity, expression, and pose as input conditions and guidance for learning a 3D-aware model. Different from the existing methods, we make the first attempt to introduce explicit 3D geometric priors to improve the faithfulness of facial shapes in reversible face anonymization.

\begin{figure*}[t]
   \begin{center}
      \includegraphics[width=1.0\linewidth]{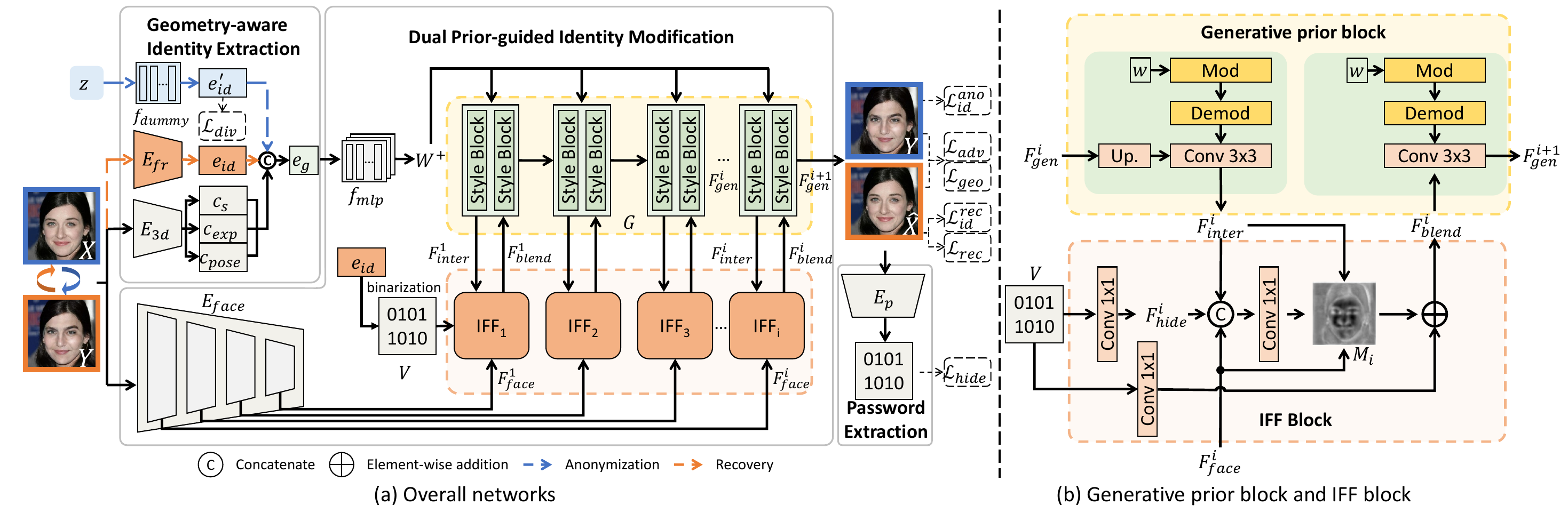}
   \end{center}
      \caption{Overview of the two-stage training pipeline for our proposed G\textsuperscript{2}Face. As the two stages share the same backbone, we showcase the main framework for simplicity, and a more detailed pipeline of each stage are illustrated in Fig.~\ref{fig:reversible}. Particularly, in the anonymization stage (blue arrow), the GIE first extracts the geometry coefficients of the input face $X$ and combines them with a randomly generated dummy ID embedding $e'_{id}$ to derive the geometry-aware identity embedding. The DPIM then generates ID-relevant features based on this embedding. These features are adaptively blended with ID-irrelevant features from the original image using the proposed identity-aware feature fusion block (IFF) in a multi-scale manner, resulting in the final anonymized face image $Y$. In the recovery stage (orange arrow), the recovered image $\hat{X}$ is generated using the geometry-aware identity embedding from the real ID embedding $e_{id}$ and the geometry coefficients of the anonymized face $Y$, along with the multi-scale feature of $Y$ by the same process as the anonymous stage. Note that during each single stage, only the corresponding ID embedding is used as the input.}
   \label{fig:structure}
   \end{figure*}
   
\subsection{Facial Attribute Editing}
The objective of the face attribute editing method is to selectively alter specific semantics of the face based on predefined conditions. Its primary goal is to disentangle the semantic attributes of the face and exclusively modify targeted features without influencing unrelated attributes~\cite{}. 
Li et al.~\cite{li2023privacy} propose a framework that obfuscates specific facial attributes while retaining facial identity. In contrast, our approach aims to conceal the identity while preserving ID-irrelevant attributes. PI-Net~\cite{chen2021perceptual} proposes a method for perceptual indistinguishability in anonymizing faces by manipulating selected semantic attributes. The PI-Net architecture generates realistic faces while preserving these chosen attributes. A3GAN~\cite{zhai2022a3gan} presents an attribute-aware anonymization network that treats face anonymization as a dual task involving semantic suppression and controllable attribute manipulation. While this method eliminates identity information, it also alters ID-irrelevant attributes, thereby leading to severe loss of data utility.

Recently, diffusion model~\cite{ho2020denoising,rombach2022high} has exhibited remarkable capabilities in the realm of generative model, particularly in face generation tasks~\cite{wang2024instantid, li2023photomaker, peng2023portraitbooth,yu2024beyond}. However, most efforts in this domain have focused on personalized ID generation, employing a global image generation process that lacks fine-grained control over various facial attributes. Due to this, diffusion-based methods struggle with accurate ID manipulation while preserving ID-irrelevant attributes. Moreover, the training and inference costs of diffusion models are higher than those of GANs, posing challenges to real-time applications. Therefore, we adopt GANs as the foundation for our framework.

\section{Proposed Method}

\subsection{Problem Definition}
We aim to anonymize an input face image by altering its identity information while precisely retaining ID-irrelevant attributes such as hairstyle, expressions, and background, and enabling faithful recovery of the original face from the anonymized counterpart simultaneously.
Formally, let $X$ and $Y$ denote the input face image and its anonymized counterpart, respectively. $e_{id}$ and $e'_{id}$ represent the corresponding ID embeddings of $X$ and $Y$, respectively. The recovered original image from $Y$ is denoted as $\hat{X}$.
In the anonymization process, given the input face image $X$ and a randomly generated dummy ID embedding $e'_{id}$, our goal is to learn a mapping $\mathcal{G}(\cdot)$ that transforms the input face into an anonymized one: $Y=\mathcal{G}(X, e'_{id})$.
In the recovery process, we aim to reconstruct the original input face $\hat{X}$ from the anonymized face $Y$ with ID embedding of the original face $e_{id}$, using the same mapping function: $\hat{X} = \mathcal{G}(Y, e_{id})$.

\subsection{Overview}
\label{sec:overview}
To achieve realistic reversible face anonymization while preserving ID-irrelevant attributes, we propose G\textsuperscript{2}Face, seamlessly combining generative and geometric priors for precise identity manipulation. The architecture, shown in Fig.~\ref{fig:structure}, consists of three modules: \textit{Geometry-aware Identity Extraction (GIE)}, \textit{Dual Prior-guided Identity Manipulation (DPIM)}, and \textit{Password Extraction (PE)}.
The GIE module includes a 3D face reconstruction model $E_{3d}$~\cite{deng2019accurate}, an ArcFace model $E_{fr}$~\cite{deng2019arcface}, and a dummy ID generator $f_{dummy}$. The DPIM comprises a face encoder $E_{face}$, a StyleGAN-based decoder $G$, and a series of identity-aware feature fusion blocks (IFFs). The PE module involves a five-layer CNN encoder $E_p$ for password embedding extraction, crucial for reversible face recovery.
Given the original image $X$, the $E_{3d}$ extracts geometric coefficients, concatenated with a dummy ID embedding $e'_{id}$ from $f_{dummy}$ to create a geometry-aware identity embedding $e_g$. This is mapped into the $\mathcal{W+}$ space to modulate the StyleGAN decoder, producing high-fidelity, identity-altered features $F^{i}_{inter}$. These features blend adaptively with ID-irrelevant features $F_{face}^i$ from the original image via IFFs, reinjected into the decoder $G$ to generate the final anonymized face image $Y$ with maintained facial details and original ID-irrelevant attributes.
For faithful face recovery, the original face's identity information $e_{id}$ is embedded into anonymized faces during IFF feature fusion. This information is extracted by the PE module for subsequent face recovery by feeding the extracted embedding and the anonymized face into the DPIM.

During each training iteration, we first generate an anonymized image $Y$ from the original image $X$ using a randomly generated dummy ID. We then recover $X$ from $Y$ using the password embedding extracted from $Y$ via the PE module $E_{p}$. In the following, we introduce each component of our model in detail.

\subsection{Geometry-aware Identity Extraction}
\label{sec:projector}

Current methods predominantly rely on ``creating" a dummy face to achieve face anonymization. However, they rarely consider keeping the geometric consistency (\eg, face shape, expression, and pose) between the anonymized face and the original face, thereby significantly undermining the data utility of the anonymized faces in downstream applications, such as landmark detection and expression recognition.
To circumvent this issue, we propose to incorporate explicit guidance into the anonymization process to ensure the geometric consistency between the anonymized face and the original face. Concretely, we introduce the 3D morphable model (3DMM)~\cite{blanz1999morphable} to provide geometric priors for synthesizing geometric-consistent and faithful anonymized images. This is the key to ensuring the data utility of face anonymization for downstream tasks.
Specifically, the face shape $\mathbf{S}$ of input image can be formulated as
\begin{equation}
   \label{eq:3dmm-shape}
  \mathbf{S} = \mathbf{S}(c_{s}, c_{exp}) = \bar{\mathbf{S}} + \mathbf{A}_{s}c_{s} + \mathbf{A}_{exp}c_{exp},
\end{equation}
where $\bar{\mathbf{S}}$ is the average face shape. $\mathbf{A}_{s}$ and $\mathbf{A}_{exp}$ are PCA bases of shape and expression, respectively. $c_{s}$ and $c_{exp}$ are the corresponding coefficient vectors for generating a 3D face. We refer the readers to the references~\cite{tu20203d,deng2019accurate} for more details.

\textbf{Dummy Identity Generation.}
To avoid the potential infringement on the identity of real individuals during anonymization, we introduce a dummy ID generation network, denoted as $f_{dummy}$, to generate dummy identity information for anonymization. $f_{dummy}$ is a two-layer MLP, which maps a random noise $z$ to a dummy identity embedding $e'_{id}$ for generating anonymized faces.

\textbf{3D Shape-aware Prior Integration.}
To ensure that the anonymized image $Y$ is geometrically consistent with the input image $X$, we propose to take advantage of the 3D shape-aware coefficients, \ie, the shape $c_s$, expression $c_{exp}$, and pose coefficients $c_{pose}$ of $X$, which are predicted by a pre-trained 3D face reconstruction model~\cite{deng2019accurate}, as the geometric prior for synthesizing anonymized image.
Thus, the geometric-aware identity embedding $e_g$ can be obtained by concatenating the identity embedding $e'_{id}$. This operation can be formulated as
\begin{equation}
e_g=\mathrm{Concat}(e'_{id}, c_s, c_{exp}, c_{pose}).
\end{equation}
Once the $e_g$ is obtained, it is ready to be sent into the DPIM for generating the anonymized face image $Y$ with randomly sampled identity and the same geometry as the input face.

\subsection{Dual Prior-guided Identity Manipulation}
For face anonymization, prior works typically learn a simple identity manipulation mapping by an encoder-decoder architecture~\cite{maximov2020ciagan, gu2020password, proencca2021uu, cao2021personalized} or solely rely on manipulating the latent code via GAN inversion~\cite{li2023riddle,barattin2023attribute}. Nevertheless, the former often tend to produce low-quality results with notable facial detail loss due to the limited generation capacity of their networks.
Although the latter explore GAN priors to achieve face anonymization with higher fidelity, they struggle with maintaining the original ID-irrelevant attributes due to the inherent inaccuracies and uncontrollability in the GAN inversion and manipulation process, thus heavily impairing the utility of anonymized faces.
To enable realistic face anonymization with data utility well preserved, we devise a dual prior-guided identity manipulation module (DPIM) that explicitly introduces generative and geometric priors to render vivid facial patterns for anonymization, while precisely retaining the original ID-irrelevant features simultaneously.
Specifically, the DPIM comprises an identity manipulation network $G$, a face encoder $E_{face}$, and a series of identity-aware feature fusion blocks (IFFs),
which are responsible for the generation of the identity-manipulated face images, the extraction of the input face features, and the adaptive fusion of the ID-relevant, ID-irrelevant features, and password embedding, respectively.

\textbf{Identity Manipulation Network.}
The identity manipulation network $G$ adopts a backbone of a pre-trained StyleGAN2~\cite{karras2020analyzing}, with its style convolution blocks modulated by the latent code $W+$ projected from the geometric-aware identity embedding $e_g$ via a latent mapper $f_{mlp}$.
Here, the $G$ is capable of generating the anonymized face image with the identity and geometry properties controlled by the geometric-aware identity embedding $e_g$.
Specifically, inspired by pSp~\cite{richardson2021encoding} that reveals different levels of latent control different grain features, we first integrate the geometric-aware identity embedding $e_g$ by three MLP projectors $f^i_{mlp}$, and then project them to $\mathcal{W}+$ space:
\begin{equation}
   w_i = f_{mlp}^i(e_g),
\end{equation}
where $i \in \{low, middle, high\}$, and these multi-level mappers can improve the quality of the generated anonymized images.

Note that for each StyleGAN2's prior block~\cite{karras2019style}, we incorporate two style convolution blocks. The intermediate feature output by the first style convolution block is denoted as follows:
\begin{equation}
   F_{inter}^{i} = Conv_{sty}(F_{gen}^i | w),
\end{equation}
where $F_{inter}^{i}$ represents the intermediate features from the first style convolution block for the $i$-th GAN block. $F_{gen}^i$ denotes the input features to that style block, and $w$ represents the latent codes modulating the StyleGAN generation process.

\begin{figure}[t!]
   \begin{center}
      \includegraphics[width=0.85\linewidth]{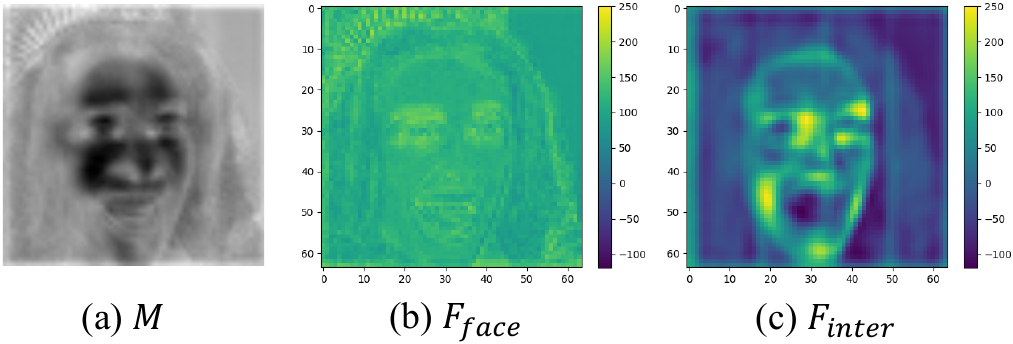}
   \end{center}
      \caption{Visualization of adaptive mask $M$, ID-irrelevant features $F_{face}$, and generative features $F_{inter}$. $M$ serves as a key indicator to guide the integration of $F_{face}$ and $F_{inter}$.}
   \label{fig:mask}
\end{figure}

\textbf{Face Encoder.}
Similar to the existing GAN inversion-based methods, simply manipulating the latent code $W+$ can inevitably contaminate the ID-irrelevant attributes of the original image.
We therefore introduce a face encoder $E_{face}$ to extract multi-scale facial features $F_{face}^i$ from the input face $X$ for providing substantial ID-irrelevant features for identity manipulation, and this process can be formulated as

\begin{equation}
   F_{face}^i = E_{face}(X, i),
\end{equation}
where $i$ is the index of the conv layer. Then, we adaptively fuse the generated feature $F_{inter}^{i}$ and multi-scale features of original $F_{face}^i$ through our proposed IFF to generate vivid anonymized faces with preserved ID-irrelevant attributes.

\textbf{Identity-aware Feature Fusion Block.} To delicately integrate the ID-irrelevant features $F_{face}^i$ and the generated ID-relevant features $F_{gen}^i$ from the $i$-th layer of $G$, we propose an identity-aware feature fusion block (IFF) that blends these two features under the guidance of the geometric-aware identity embedding $e_g$.
Concurrently, following the same spirit from steganography~\cite{zhang2019steganogan,feng2014secure,tang2019cnn}, we embed the identity information $e_{id}$ of the original image $X$ into the anonymized face $Y$ in the anonymization process to facilitate subsequent reversible recovery. Considering that $e_{id}$ is a 512-dimension vector consisting of 512 floating-point values.
During the training phase, we extract $e_{id}$ from $X$ and convert it into a binary vector with 16,384 (512$\times$32) bits. Subsequently, we reshape this binary vector into a password embedding denoted as $V \in \{0,1\}^{D \times H \times W}$, where $H$ and $W$ represents the height and width of features in the $i$-th conv layer, and $D$ signifies the depth of $V$, which can be computed as $16384/HW$.
Once the password embedding $V$ is obtained, it is then fed into a $1\times1$ convolutional layer to produce the hiding features $F_{hide}^i$ as follows:
\begin{equation}
   F_{hide}^i = Conv_{1 \times 1}(V),
\end{equation}
where $i$ is the index of the convolutional layer.

\begin{figure}[t]
   \begin{center}
      \includegraphics[width=0.9\linewidth]{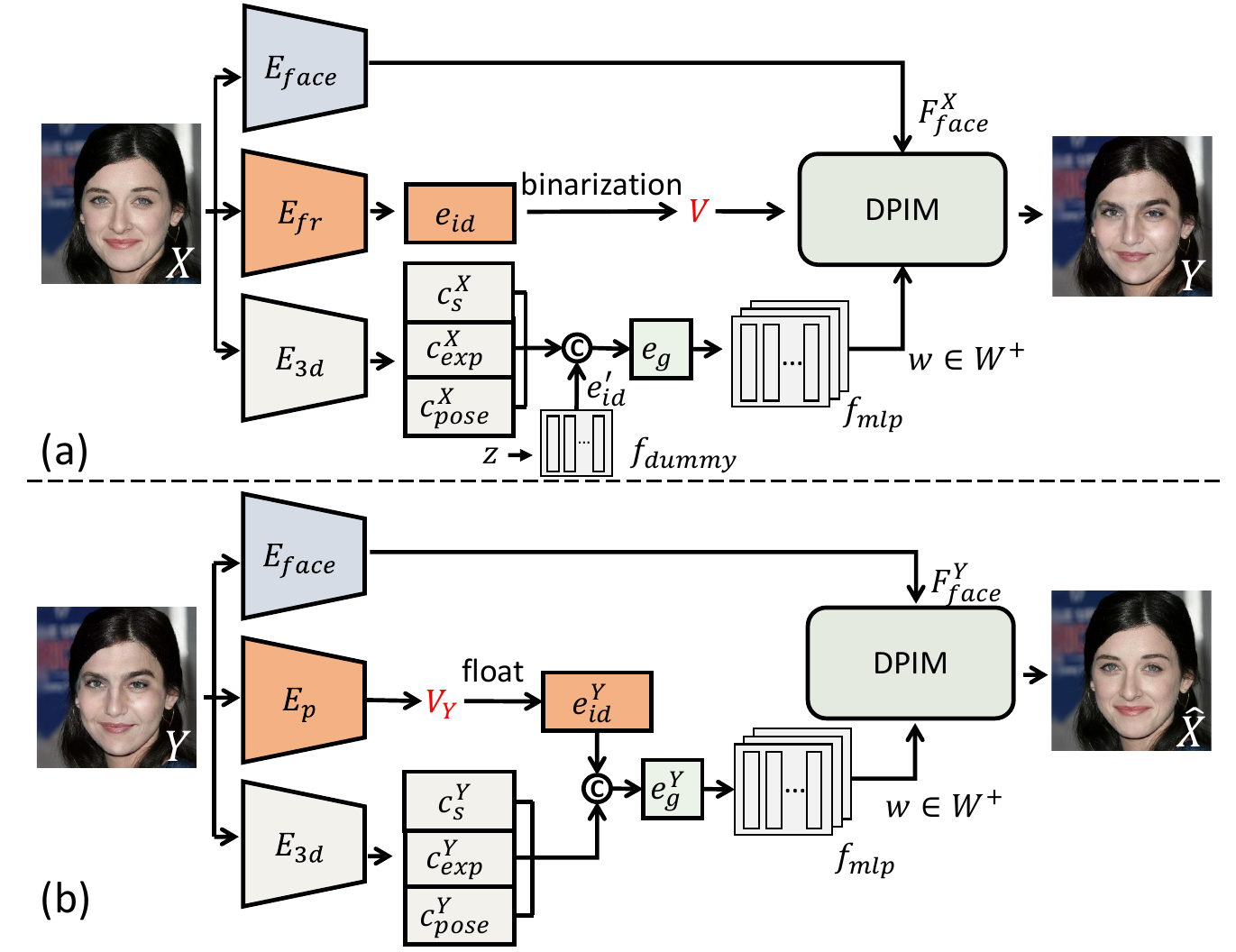}
   \end{center}
      \caption{The inference pipeline for our proposed reversible face anonymization. (a) Face anonymization. (b) Face recovery.}
   \label{fig:reversible}
   \end{figure}

As shown in Fig.~\ref{fig:structure}(b), each IFF takes as input the $F_{face}^i$, $F_{inter}^{i}$, and $F_{hide}$.
Firstly, the $F_{face}^i$, $F_{inter}^{i}$, and $F_{hide}^i$ are integrated by applying a 1$\times$1 convolution, yielding an adaptive mask $M_i$ via a Sigmoid operation:
\begin{equation}
M_{i} = Conv_{1 \times 1}(F_{face}^i, F_{inter}^{i}, F_{hide}^i).
\end{equation}
Note that here the mask $M_i$ can adaptively captures the ID-irrelevant pixels, which is fully learned with indirect supervision from the final anonymized/recovered results. It serves as a key indicator to guide the integration of generated features and original ID-irrelevant features, thereby enabling precise high-fidelity identity manipulation.
With the guidance of $M_i$, the feature blending process can formulated as follows:
\begin{equation}
F_{blend} ^ i = M_i \cdot F_{face}^i + (1-M_i) \cdot F_{inter}^{i} + Conv_{1 \times 1}(V).
\end{equation}
Here, $F_{blend}^i$ represents the blended features, which are then fed back into the subsequent convolutional layers of $G$ for feature generation.

We present the visualization of the mask $M$, original face features $F_{face}$, and generative features $F_{inter}$ in Fig.~\ref{fig:mask}.
The results show that the adaptive mask indeed effectively captures the ID-irrelevant regions, particularly encompassing background and hairstyle regions. Simultaneously, the generative model generates anonymized key attributes related to ID-relevant facial characteristics. Our proposed adaptive feature fusion block adeptly merges these two features, leading to a visually compelling anonymized image with preserved ID-irrelevant attributes.

\subsection{Reversible Face Anonymization Process}

To enable reversible recovery, we initiate the process by training a Password Extractor, denoted as $E_p$, which consists of multiple convolutional layers designed specifically to extract the binary password. During the training phase, we first convert the ID embedding $e_{id}$ into its binary format $V$, \ie, $V = \text{binary}(e_{id})$. This binary password $V$ is then embedded into the generated image using the IFF block. Subsequently, we leverage the Password Extractor $E_p$ to predict $V$ and optimize it with the Information Hiding Loss in Sec.~\ref{sec:loss}.

Fig. 4 illustrates the inference pipeline for reversible face anonymization.
At the anonymous stage of the inference process (See Fig.~\ref{fig:reversible}(a)), we first transform the original ID embedding $e_{id}$ from its floating-point representation $e_{id}$ to the binary format $V$ and embed it as a password by our IFF blocks within the anonymized image $Y$, then anonymize the image by the random generated ID latent $e'_{id}$. Moving forward to the reversible recovery phase (See Fig.~\ref{fig:reversible}(b)), the binary password embedded within the anonymous image is extracted using the Password Extractor $E_p$, resulting in $V_{Y} = E_{p}(Y)$, and then converts it back into its corresponding floating-point identity embedding $e^Y_{id}$. Afterwards, the 3DMM shape coefficient $c_{s}^{Y}$, expression coefficient $c_{exp}^{Y}$, and pose coefficient $c_{pose}^{Y}$ extracted from $Y$ are concatenated together with $e^Y_{id}$, yielding the final geometric-aware identity embedding $e^Y_g$. Then, $e^Y_g$ is projected into the $\mathcal{W}+$ space via the mapper network $f_{mlp}^i$:

\begin{equation}
   w_i^{rec} = f_{mlp}^i(e^Y_g).
\end{equation}

Here, $w_i^{rec}$ represents the recovery latent in the $\mathcal{W}+$ space and $i \in \{low, middle, high\}$. Finally, the original image $\hat{X}$ can be recovered by passing the facial features extracted by $E_{face}$ from $Y$ and the projected vector $w_i^{rec} \in \mathcal{W}+$ through $G$:

\begin{equation}
\hat{X} = G(E_{face}(Y), w_i^{rec}, V).
\end{equation}

Note that input $V$ during the recovery phase just acts as a placeholder and there is no need for information hiding.

\subsection{Loss Function}
\label{sec:loss}
To enable efficient learning of the proposed model, we adopt a two-stage training process. Firstly, we generate an anonymized image $Y$ from the original image $X$ using a randomly generated identity vector. Subsequently, we extract the password embedded in $Y$ and recover the original image $\hat{X}$.
To maintain consistency and simplicity in our loss functions across both stages, we introduce the following notation: $X_{in}$ to present the input image to the model. It corresponds to the original face $X$ in the anonymization stage and the anonymized face $Y$ in the recovery stage.
Conversely, $X_{out}$ signifies the output image of the model, which refers to the anonymized image $Y$ in the anonymization stage, and the recovered image $\hat{X}$ in the recovery stage. We adopt the reconstruction loss, adversarial loss, identity loss, diversity loss, geometry loss, and information hiding loss to govern our end-to-end training.

\textbf{Reconstruction Loss.}
Since we have no paired data for the anonymization-stage training, we apply a reconstruction loss on the output image $X_{out}$ in the recovery stage to enforce its content consistency with the original input face $X$.

\begin{equation}
   \begin{aligned}
   \mathcal{L}_{rec}(X_{out}, X) = \Vert X_{out}-X \Vert_1 + & \operatorname{LPIPS}(X_{out}, X),
   \end{aligned}
\end{equation}
where $\Vert \cdot \Vert_1$ represents the L1-norm, and $\operatorname{LPIPS}(\cdot, \cdot)$ denote the perceptual similarity~\cite{zhang2018unreasonable}.

\textbf{Adversarial Loss.}
Following StyleGAN2~\cite{karras2020analyzing}, we also adopt an adversarial loss to improve the fidelity of generated images $X_{out}$:
\begin{equation}
   \mathcal{L}_{adv}(X_{out}) = \mathbb{E}_{X_{out}} \log(1+\exp(-D(X_{out}))),
\end{equation}
where $D$ is the discriminator that shares a similar structure with StyleGAN2's discriminator~\cite{karras2020analyzing}.

\textbf{Identity Loss.}
In the anonymization stage, our goal is to maximize the divergence of the identity (ID) of the synthesized face from the original ID. Additionally, the generated image ID is expected to align with the dummy ID latent $e'_{id}$ in the anonymization stage. Hence, we adopt an identity loss in the anonymization stage, which can be formulated as
\begin{equation}
   \begin{aligned}
   \mathcal{L}_{id}^{ano}(Y, e_{id}, e'_{id}) = & 1-\cos(e'_{id}, E_{fr}(Y)) \\
                                                & + \max(0, \cos(e_{id}, E_{fr}(Y))),
   \end{aligned}
\end{equation}
where $e_{id}$ stands for the original ID embedding of input image $X$. $E_{fr}(\cdot)$ signifies the identity embedding extracted by a pre-trained ArcFace~\cite{deng2019arcface}.

During the recovery stage, the aim is to align the ID of the recovered image $\hat{X}$ closely with the original $e_{id}$. We therefore adopt an identity loss in the reconstruction phase computed as follows:
\begin{equation}
   \mathcal{L}_{id}^{rec}(X, e_{id}) = 1-\cos(e_{id}, E_{fr}(\hat{X})).
\end{equation}

\textbf{Diversity Loss.}
To improve the diversity of the generated anonymized faces, we employ a diversity loss that encourages the diversity of the generated identity latent space.
Specifically, for a mini-batch training data with $N_s$ samples, we generate $N_s$ corresponding dummy ID embeddings to anonymize these samples. The similarity between any two ID latents can be represented by a similarity matrix $\mathcal{M}$ with a size of $N_s \times N_s$. The cosine similarity between two elements at the index of $i$ and $j$ can be calculated as:

\begin{equation}
   \label{eq:div}
   \begin{aligned}
      \mathcal{M}_{i,j} = \max(0, \cos (e_{id}^{'i}, e_{id}^{'j})),
   \end{aligned}
\end{equation}
where $\cos(\cdot, \cdot)$ refers to the cosine similarity. The diversity loss of the dummy IDs can be then computed as:
\begin{equation}
   \label{eq:l-div}
   \begin{aligned}
      \mathcal{L}_{div} = \frac{1}{n(n-1)}  \sum (\mathbf{(1-I)} \mathcal{M}),
   \end{aligned}
\end{equation}
where $\mathbf{1}$ is a all-one matrix and $\mathbf{I}$ is an identity diagonal matrix.
By minimizing the sum of cosine similarities between each pair of dummy IDs, we can effectively improve the diversity of the generated anonymized faces.

\textbf{Geometry Loss.} Ensuring geometric consistency (\eg, face shape and expression) between the output and input images is crucial, particularly for maintaining the data utility of anonymized faces in downstream facial analysis tasks. To this end, we employ a geometry loss on the output faces of both stages. Specifically, a pre-trained 3D face reconstruction model~\cite{deng2019accurate}, denoted as $E_{3d}$, is used to predict the 3D face shapes of the input and output faces (Eq.~\eqref{eq:3dmm-shape}). Here, the 3D shape of the original face $X$ and the the output face $X_{out}$ are denoted as $\mathbf{S}_{X}$ and $\mathbf{S}_{X_{out}}$, respectively. The color of the input face $X$ is represented as $\mathbf{C}_{X}$, and the color of $X_{out}$ as $\mathbf{C}_{X_{out}}$ (Eq.~\eqref{eq:3dmm-color}).

To start with, we introduce a mesh loss to ensure the 3D shape consistency between $\mathbf{S}_{X_{out}}$ and $\mathbf{S}_X$ and the color consistency between $\mathbf{C}_{X_{out}}$ and $\mathbf{C}_X$ at all vertices:
\begin{equation}
   \mathcal{L}_{mesh}(X, X_{out}) = ||\mathbf{S}_X - \mathbf{S}_{X_{out}} ||_2^2 + || \mathbf{C}_X - \mathbf{C}_{X_{out}} ||_2^2.
\end{equation}
Here, the color $\mathbf{C}$ at each vertex $i$ is defined as follows:
\begin{equation}
    \label{eq:3dmm-color}
   \mathbf{C}(i) = \mathbf{c}(\mathbf{n}_i, \gamma) = \sum_{b=1}^{27} \gamma_b \Phi_b (\mathbf{n}_i),
\end{equation}
where $\mathbf{n}_i$ denotes the surface normal at vertex $i$ and $\Phi_b$ is the Spherical Harmonics(SH) basis function~\cite{tu20203d,deng2019accurate}.

Then, to ensure expression consistency, we also adopt a 3D landmark loss, which is formulated as
\begin{equation}
\mathcal{L}_{lm}(X, X_{out}) = \frac{1}{N}\sum_{i=1}^{N} || \operatorname{lm}(X)_i - \operatorname{lm}(X_{out})_i ||_2^2,
\end{equation}
where $\operatorname{lm}(\cdot)$ denotes the facial landmarks predicted by the 3D face model~\cite{deng2019accurate}. $N$ denotes the number of landmarks, which is set to 68.

The overall geometry loss can be calculated as
\begin{equation}
   \mathcal{L}_{geo}(X, X_{out}) = \mathcal{L}_{mesh}(X, X_{out}) + \lambda_{lm} \mathcal{L}_{lm}(X, X_{out}),
\end{equation}
where $\lambda_{lm}$ is the weighting parameter, which is empirically set as $0.01$.

\textbf{Information Hiding Loss.}
To effectively extract the password from the generated face image, we apply an information hiding loss formulated as:

\begin{equation}
\begin{aligned}
\mathcal{L}_{hide} = \mathcal{L}_{\operatorname{BCE}}(E_p(X_{out}+\epsilon), V).
\end{aligned}
\end{equation}
Here, $\mathcal{L}_{\operatorname{BCE}}$ represents the binary cross-entropy loss, and $\epsilon$ refers to a small Gaussian noise for enhancing the robustness of the information embedding process. $E_p(\cdot)$ denotes the password extracted by the password extractor.

\renewcommand{\algorithmicrequire}{\textbf{Input:}}  
\renewcommand{\algorithmicensure}{\textbf{Output:}}

\begin{algorithm}[t]
    \caption{Training strategy of G\textsuperscript{2}Face.}
    \label{algo}
    \begin{algorithmic}[1]
    \Require Input image $X$, iteration steps $steps$, Arcface model $E_{fr}$, D3DFR model $E_{3d}$.
    \Ensure Dummy ID network $f_{dummy}$, MLP projectors $f_{mlp}$, face encoder $E_{face}$, PE module $E_p$, Decoder $G$.

    \For{$i = 1$ \textbf{to} $steps$}
        \State Random sample $z$.  \Comment{Anonymization stage}
        \State $e_{id}=E_{fr}(X)$
        \State Convert $e_{id}$ to its binary representation $V$.
        \State $c_s, c_{exp}, c_{pose}=E_{3d}(X)$   \Comment{Geo. prior}
        \State $e'_{id}=f_{dummy}(z)$         \Comment{Dummy ID} 
        \State $e_g=\operatorname{Concat}(e'_{id}, c_s, c_{exp}, c_{pose})$
        \State $Y=G(E_{face}(X), f_{mlp}(e_g), V)$
        \State $V_Y = E_p(Y)$                \Comment{Password extraction}
        \State Convert $V_Y$ to its floating representation $e_{id}^Y$.
        \State Compute anoymization loss by Eq.~\eqref{eq: total_ano}.
    
        \State $c_s^Y, c_{exp}^Y, c_{pose}^Y=E_{3d}(Y)$ \Comment{Recovery stage}
        \State $e_g^Y=\operatorname{Concat}(e^Y_{id}, c_s^Y, c_{exp}^Y, c_{pose}^Y)$
        \State $\hat{X}=G(E_{face}(Y), f_{mlp}(e_g^Y), V)$ 
        \State $V = E_p(\hat{X})$               
        \State Compute recovery loss by Eq.~\eqref{eq: total_rec}.
        \State Update model.
    \EndFor

    \end{algorithmic}
    \end{algorithm}

\textbf{Total Loss.}
Therefore, the total loss of our two-stage training is as follows:
\begin{equation}
   \begin{aligned}
      \mathcal{L}_{ano}(X, Y) = & \lambda_{div}\mathcal{L}_{div} + \lambda_{adv} \mathcal{L}_{adv} + \lambda_{id} \mathcal{L}^{ano}_{id}  \\
                                & + \lambda_{geo}\mathcal{L}_{geo}  + \lambda_{hide} \mathcal{L}_{hide} ,
   \end{aligned}
   \label{eq: total_ano}
\end{equation}
and
\begin{equation}
   \begin{aligned}
      \mathcal{L}_{rec}(Y, \hat{X}) = & \lambda_{rec}\mathcal{L}_{rec} + \lambda_{adv}\mathcal{L}_{adv} + \lambda_{id}\mathcal{L}^{rec}_{id} \\
                                      & + \lambda_{geo}\mathcal{L}_{geo} + \lambda_{hide} \mathcal{L}_{hide} ,
   \end{aligned}
   \label{eq: total_rec}
\end{equation}
where $\lambda_{div}$, $\lambda_{rec}$, $\lambda_{adv}$, $\lambda_{id}$, $\lambda_{geo}$, and $\lambda_{hide}$ are weighting parameters to balance the above loss items, which are set as 1, 1, 1, 1, 1, and 10, respectively.
We show the training process of our model in Algorithm~\ref{algo}.

\section{Experiments}

\subsection{Settings}
\label{sec:setting}
\textbf{Experimental Settings.}
Our model is implemented using the PyTorch framework. We adopt a pre-trained StyleGAN2\cite{karras2020analyzing} as the backbone of the DPIM, which offers rich generative priors. The $f_{mlp}$ is a four-layer MLP.
The framework is trained using Adam~\cite{kingma2014adam} ($\beta_1=0.90$, $\beta_2=0.99$) with a learning rate of 0.0001 for the StyleGAN2 block and 0.001 for the other trainable parts. The batch size is set as 8.
The entire model is trained on 4 NVIDIA RTX4090 GPUs.

\textbf{Datasets.}
We train our model on the FFHQ dataset~\cite{karras2019style}, which comprises 70,000 face images. For evaluation, we choose the CelebA-HQ dataset~\cite{karras2018progressive}, which consists of 30,000 images. We also evaluate our method on the LFW\cite{huang2008labeled} dataset to validate its generalization capability. All the images are aligned and cropped to the size of 256 $\times$ 256.

\textbf{Baselines.} For the anonymization experiments, we compare our method with several SOTA face anonymization approaches both qualitatively and quantitatively, including CIAGAN~\cite{maximov2020ciagan}, FIT~\cite{gu2020password}, RiDDLE~\cite{li2023riddle}, and FALCO~\cite{barattin2023attribute}. 
For the recovery experiments, we included the only two open-source methods that support reversible face anonymization, allowing for both qualitative and quantitative comparisons. These methods are FIT~\cite{gu2020password} and RiDDLE~\cite{li2023riddle}.

\subsection{Anonymization Comparison with SOTA Methods}
\label{sec:anonymization}
\begin{figure*}[htbp]
    \centering
    \subfloat[Input]{
        \begin{minipage}{0.1\linewidth}
            \includegraphics[height=\linewidth]{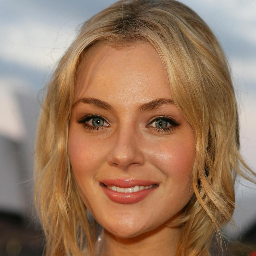} \\
            \includegraphics[height=\linewidth]{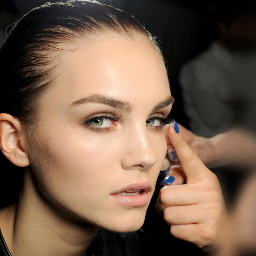} \\
            \includegraphics[height=\linewidth]{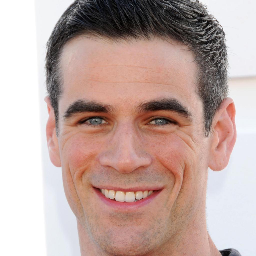} \\
            \includegraphics[height=\linewidth]{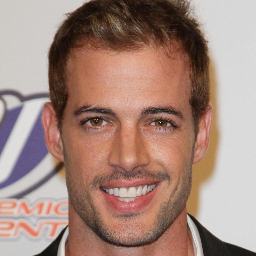} \\
            \includegraphics[height=\linewidth]{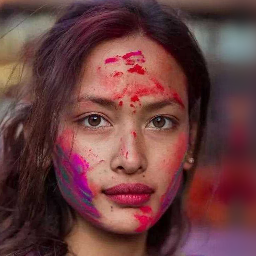} \\
        \end{minipage}
    }
    \hspace{-6pt}
    \subfloat[CIAGAN]{
        \begin{minipage}{0.1\linewidth}
            \includegraphics[height=\linewidth]{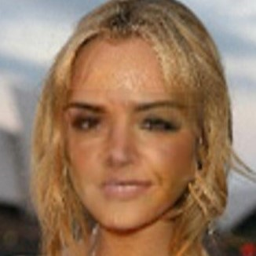} \\
            \includegraphics[height=\linewidth]{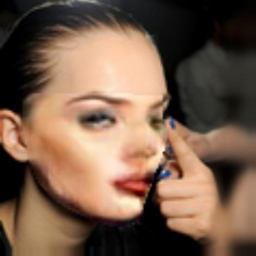} \\
            \includegraphics[height=\linewidth]{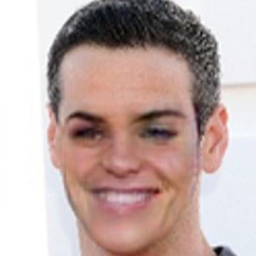} \\
            \includegraphics[height=\linewidth]{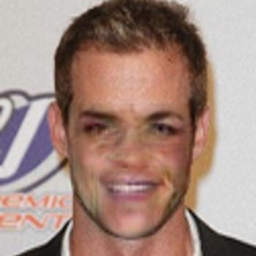} \\
            \includegraphics[height=\linewidth]{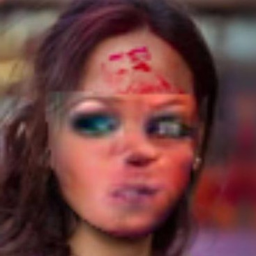} \\
        \end{minipage}
    }
    \hspace{-12pt}
    \subfloat[FIT]{
        \begin{minipage}{0.1\linewidth}
            \includegraphics[height=\linewidth]{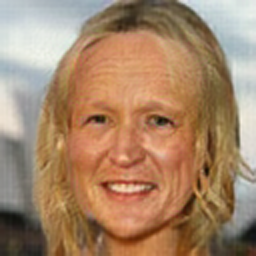} \\
            \includegraphics[height=\linewidth]{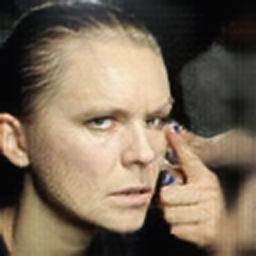} \\
            \includegraphics[height=\linewidth]{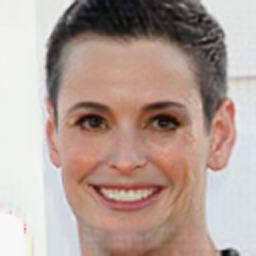} \\
            \includegraphics[height=\linewidth]{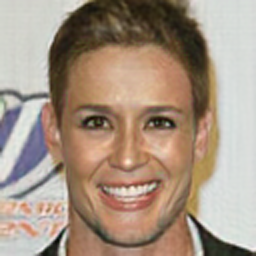} \\
            \includegraphics[height=\linewidth]{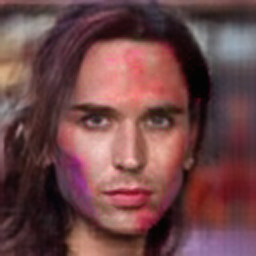} \\
        \end{minipage}
    }
    \hspace{-12pt}
    \subfloat[RiDDLE]{
        \begin{minipage}{0.1\linewidth}
            \includegraphics[height=\linewidth]{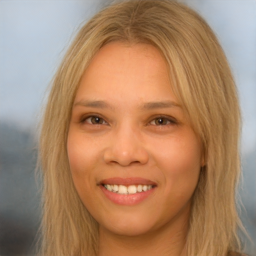} \\
            \includegraphics[height=\linewidth]{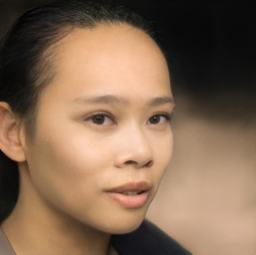} \\
            \includegraphics[height=\linewidth]{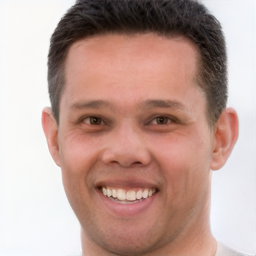} \\
            \includegraphics[height=\linewidth]{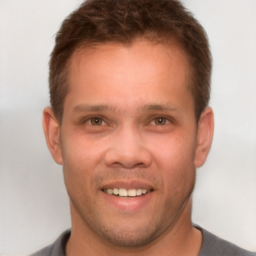} \\
            \includegraphics[height=\linewidth]{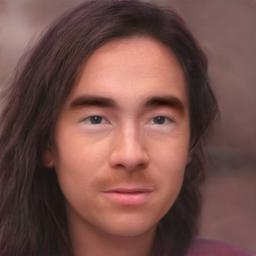} \\
        \end{minipage}
    }
    \hspace{-12pt}
    \subfloat[FALCO]{
        \begin{minipage}{0.1\linewidth}
            \includegraphics[height=\linewidth]{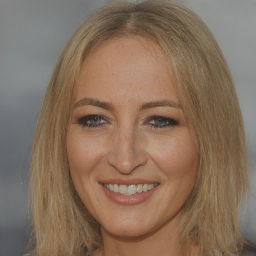} \\
            \includegraphics[height=\linewidth]{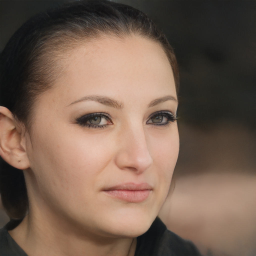} \\
            \includegraphics[height=\linewidth]{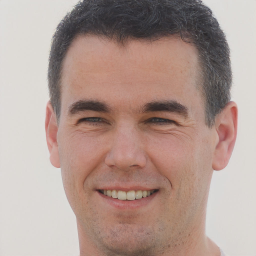} \\
            \includegraphics[height=\linewidth]{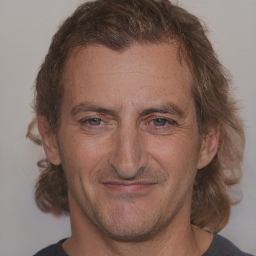} \\
            \includegraphics[height=\linewidth]{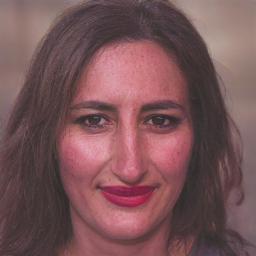} \\
        \end{minipage}
    }
    \hspace{-12pt}
    \subfloat[Ours]{
        \begin{minipage}{0.1\linewidth}
            \includegraphics[height=\linewidth]{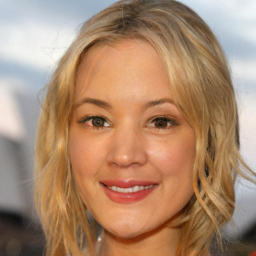} \\
            \includegraphics[height=\linewidth]{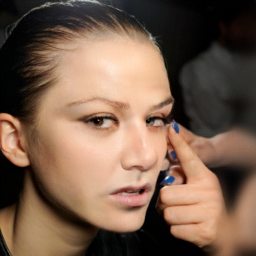} \\
            \includegraphics[height=\linewidth]{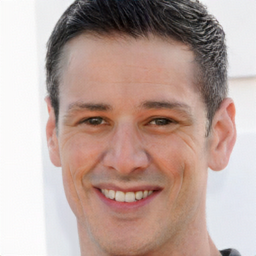} \\
            \includegraphics[height=\linewidth]{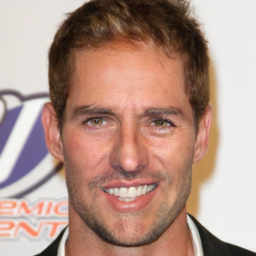} \\
            \includegraphics[height=\linewidth]{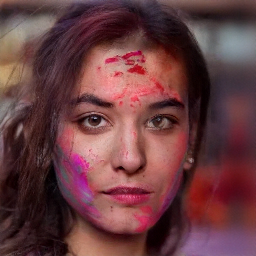} \\
        \end{minipage}
    }
    \hspace{-6pt}
    \subfloat[FIT-R]{
        \begin{minipage}{0.1\linewidth}
            \includegraphics[height=\linewidth]{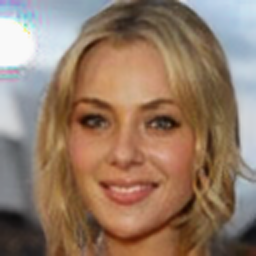} \\
            \includegraphics[height=\linewidth]{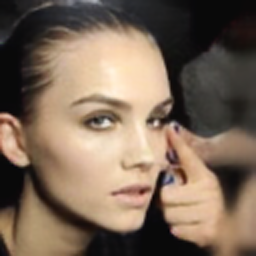} \\
            \includegraphics[height=\linewidth]{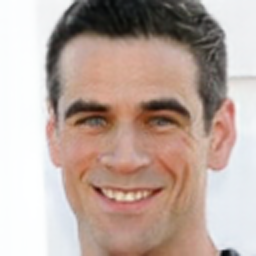} \\
            \includegraphics[height=\linewidth]{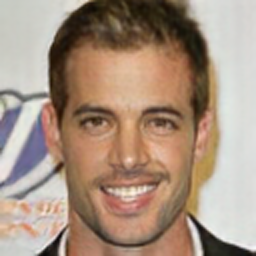} \\
            \includegraphics[height=\linewidth]{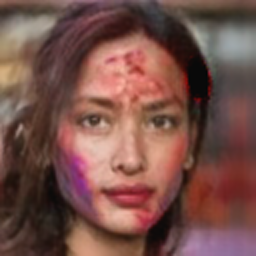} \\
        \end{minipage}
    }
    \hspace{-12pt}
    \subfloat[RiDDLE-R]{
        \begin{minipage}{0.1\linewidth}
            \includegraphics[height=\linewidth]{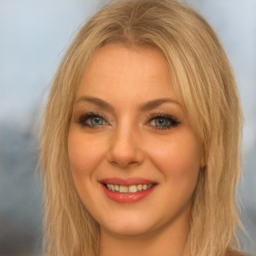} \\
            \includegraphics[height=\linewidth]{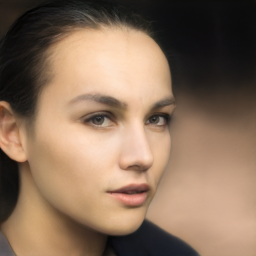} \\
            \includegraphics[height=\linewidth]{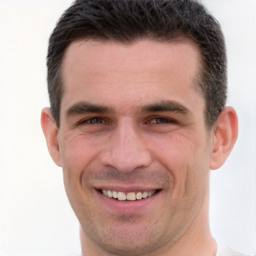} \\
            \includegraphics[height=\linewidth]{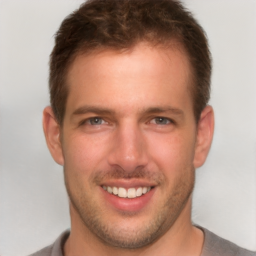} \\
            \includegraphics[height=\linewidth]{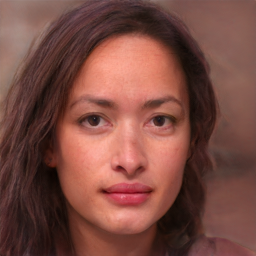} \\
        \end{minipage}
    }
    \hspace{-12pt}
    \subfloat[Ours-R]{
        \begin{minipage}{0.1\linewidth}
            \includegraphics[height=\linewidth]{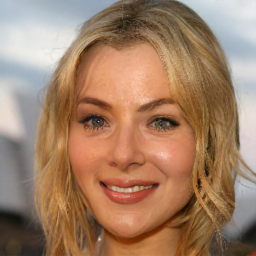} \\
            \includegraphics[height=\linewidth]{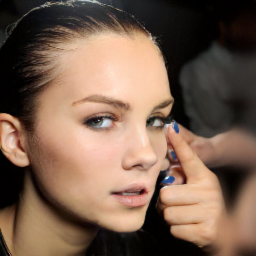} \\
            \includegraphics[height=\linewidth]{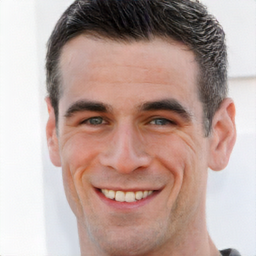} \\
            \includegraphics[height=\linewidth]{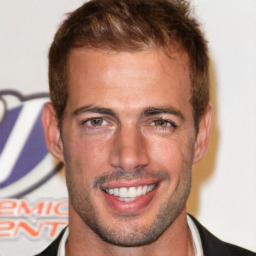} \\
            \includegraphics[height=\linewidth]{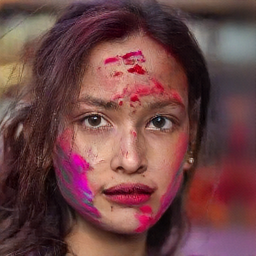} \\
        \end{minipage}
    }
    \caption{
        Qualitative comparison of face anonymization and recovery among different methods on the CelebA-HQ dataset~\cite{karras2018progressive}. (a)-(f) are the original face images, the anonymization results of CIAGAN~\cite{maximov2020ciagan}, FIT~\cite{gu2020password}, RiDDLE~\cite{li2023riddle}, FALCO~\cite{barattin2023attribute}, and our method, respectively. (g) to (i) are the recovery results of FIT~\cite{gu2020password}, RiDDLE~\cite{li2023riddle}, and our G\textsuperscript{2}Face, respectively. Note that only FIT~\cite{gu2020password} and RiDDLE~\cite{li2023riddle} support reversible face anonymization.}
     \label{fig:visual_comparison}
\end{figure*}

\textbf{Qualitative Comparison.}
To demonstrate the effectiveness of our method in face anonymization, we first compare our method with four SOTA methods qualitatively on the CelebA-HQ dataset. Figs.~\ref{fig:visual_comparison}(b)-(f) present the face anonymization visualization results of different methods. As is observed, although CIAGAN~\cite{maximov2020ciagan} and FIT~\cite{gu2020password} can alter the identity of the input faces to some extent, they suffer from notable face distortions (Fig.~\ref{fig:visual_comparison}(b)) or obscure facial textures (Fig.~\ref{fig:visual_comparison}(c)) due to limited capacity of their naive encoder-decoder based network. On the other hand, despite that RiDDLE~\cite{li2023riddle} and FALCO~\cite{barattin2023attribute} can produce more visually plausible results with rich facial details, they fail to preserve the original id-irrelevant attributes (\eg, hairstyles, background, expressions) of the input faces as they perform anonymization through simply manipulating the latent code after GAN inversion (Figs.~\ref{fig:visual_comparison}(d) and (e)).
In contrast, our method not only generates realistic anonymized faces with faithful facial details and shapes but also perfectly preserves the original id-irrelevant attributes (Fig.~\ref{fig:visual_comparison}(f)).
This is mainly attributed to our delicate exploitation of generative and geometric priors with precise identity awareness.
It is also worth noting that, our method can perform satisfactory face anonymization even when the input face contains heavy occlusions compared with other methods (the fifth row of Fig.~\ref{fig:visual_comparison}). This also indicates the effectiveness of our proposed IFF module in precisely integrating the generated identity features and the original id-irrelevant attributes.

\begin{table}[t]
   \centering
   \caption{The true acceptance rate of face anonymization on the LFW~\cite{huang2008labeled} dataset. Bold and underlined indicate the best and second-best results, respectively.}
   \setlength{\tabcolsep}{3pt}
   \begin{tabular}{@{}ccccc@{}}
   \toprule
  Method & \begin{tabular}[c]{@{}c@{}}FaceNet $\downarrow$\\ (VGGFace2)\end{tabular} & \begin{tabular}[c]{@{}c@{}}FaceNet $\downarrow$\\ (CASIA)\end{tabular} & \begin{tabular}[c]{@{}c@{}}ArcFace $\downarrow$\\ (MS1MV3)\end{tabular} & \begin{tabular}[c]{@{}c@{}}AdaFace $\downarrow$ \\ (WebFace12M)\end{tabular} \\ \midrule
   Original                           & 0.986 ± 0.010                      & 0.965 ± 0.016                   & 0.987 ± 0.015                              & 0.989 ± 0.014                                \\
   Gafni\cite{gafni2019live}          & 0.038 ± 0.015                      & 0.035 ± 0.011                   & 0.025 ± 0.014                              & 0.037 ± 0.018                                \\
   CIAGAN\cite{maximov2020ciagan}     & 0.034 ± 0.016                      & \underline{0.019 ± 0.008}       & 0.021 ± 0.010                              & 0.027 ± 0.011 \\
   FIT\cite{gu2020password}           & 0.031 ± 0.017                      & 0.033 ± 0.008                   & 0.027 ± 0.018                              & 0.033 ± 0.021 \\
   RiDDLE\cite{li2023riddle}          & \underline{0.016 ± 0.003}          & 0.032 ± 0.006                   & \underline{0.011 ± 0.008}                  & 0.025 ± 0.012 \\
   FALCO\cite{barattin2023attribute}  & 0.016 ± 0.005                      & 0.021 ± 0.007                   & 0.015 ± 0.009                              & \underline{0.022 ± 0.011} \\ \midrule
   Ours                               & \textbf{0.010 ± 0.003}             & \textbf{0.009 ± 0.004}          & \textbf{0.007 ± 0.005}                     & \textbf{0.019 ± 0.010} \\ \bottomrule
   \end{tabular}
   \label{tab:lfw}
   \end{table}

\textbf{Quantitative Comparison.}
We also conduct a quantitative comparison of face anonymization between our method and the existing methods on the LFW dataset~\cite{huang2008labeled}.
Here, following the same setting as~\cite{maximov2020ciagan, gu2020password, li2023riddle}, we choose the true acceptance rate: the ratio of true positives for a maximum 0.001 ratio of false positives, as our metric. In our approach, we anonymize the second face image within each positive pair from the LFW dataset. We then evaluate the true acceptance rate of all pairs using two FaceNet models~\cite{schroff2015facenet}, trained on VGGFace2~\cite{cao2018vggface2} and CASIA-WebFace~\cite{yi2014learning}, as well as ArcFace~\cite{deng2019arcface}, trained on MS1MV3~\cite{deng2019lightweight}, and AdaFace~\cite{kim2022adaface}, trained on WebFace12M~\cite{zhu2021webface260m}, for face recognition assessment.
As shown in Table~\ref{tab:lfw}, our method achieves the lowest true acceptance rate among all the compared methods, indicating its superior anonymization performance.

\begin{table}[t]
   \centering
   \caption{Cosine similarity between the original and anonymized faces on the CelebA-HQ dataset. Bold and underlined indicate the best and second-best results, respectively.}
   \setlength{\tabcolsep}{3pt}{
   \begin{tabular}{@{}ccccc@{}}
   \toprule
   Method & \begin{tabular}[c]{@{}c@{}}  FaceNet  $\downarrow$ \\ (VGGFace2) \end{tabular} & \begin{tabular}[c]{@{}c@{}}  ArcFace  $\downarrow$ \\ (MS1MV3) \end{tabular} & \begin{tabular}[c]{@{}c@{}}  AdaFace  $\downarrow$ \\ (WebFace12M) \end{tabular}    \\ \midrule
   CIAGAN\cite{maximov2020ciagan}     & 0.2330 ± 0.0039               & 0.2251 ± 0.0092                  &  0.2481 ± 0.0038 \\
   FIT\cite{gu2020password}           & 0.2169 ± 0.0024               & 0.2267 ± 0.0065                  &  0.1977 ± 0.0045   \\
   RiDDLE\cite{li2023riddle}          & \underline{0.1942 ± 0.0028}   & \underline{0.1324 ± 0.0080}      &  \underline{0.1491 ± 0.0051}   \\
   FALCO\cite{barattin2023attribute}  & 0.2269 ± 0.0019               & 0.1620 ± 0.0075                  &  0.1713 ± 0.0027                  \\ \midrule
   \textbf{Ours}                               & \textbf{0.1757 ± 0.0012}      & \textbf{0.1055 ± 0.0006} & \textbf{0.1296 ± 0.0014}       \\ \bottomrule
   \end{tabular}}
   \label{tab:celebahq}
\end{table}

For a more comprehensive comparison, we also conduct a quantitative evaluation on the CelebA-HQ dataset. Since there is no paired identity label in CelebA-HQ, we apply our anonymization technique to each sample in the testing set and calculate the cosine similarity between the identity embeddings of the original faces and the anonymized faces. To ensure a convincing comparison, we adopt three different models for identity embedding extraction, including the FaceNet~\cite{schroff2015facenet} trained on VGGFace2~\cite{cao2018vggface2}, the ArcFace~\cite{deng2019arcface} trained on MS1MV3~\cite{deng2019lightweight}, and the AdaFace~\cite{kim2022adaface} trained on WebFace12M~\cite{zhu2021webface260m}, respectively.
As revealed in Table~\ref{tab:celebahq}, our method can produce anonymized results that share less identity similarity to the original faces, compared with other methods. This further demonstrates the superior anonymization performance of our method for different datasets.

\begin{table*}[t]
   \centering
   \caption{Utility maintenance comparison of different methods on the CelebA-HQ dataset. For a more comprehensive evaluation, we choose two different models for each downstream task. Bold and underlined indicate the best and second-best results, respectively.}
   \setlength{\tabcolsep}{3mm}{
   \begin{tabular}{@{}ccccccc@{}}
   \toprule
   \multicolumn{2}{c}{Method}                                                                                                                                                    & CIAGAN\cite{maximov2020ciagan}  & FIT\cite{gu2020password}     & RiDDLE\cite{li2023riddle}          & FALCO\cite{barattin2023attribute}           & Ours             \\ \midrule
   \multicolumn{2}{c|}{FID$\downarrow$ }                                                                                                                                         & 29.0300 ± 2.1812   & 20.6280 ± 1.9214  & 15.4870 ± 3.9812           & \underline{14.1120 ± 1.2356}          & \textbf{7.0412 ± 0.0312}    \\ \midrule
   \multicolumn{1}{c|}{\multirow{2}{*}{\begin{tabular}[c]{@{}c@{}}Face\\ Detection$\uparrow$\end{tabular}}}                                       & \multicolumn{1}{c|}{MTCNN}   & 0.9955 ± 0.0002    & 0.9983 ± 0.0001     & \textbf{0.9998 ± 0.0001} & \underline{0.9990 ± 0.0002}         & 0.9883 ± 0.0001          \\
   \multicolumn{1}{c|}{}                                                                                                                          & \multicolumn{1}{c|}{Dlib}    & 0.9755 ± 0.0003    & 0.9775 ± 0.0002     & \underline{0.9810 ± 0.0001}         & \textbf{0.9822 ± 0.0002} & 0.9779 ± 0.0004          \\ \midrule
   \multicolumn{1}{c|}{\multirow{2}{*}{\begin{tabular}[c]{@{}c@{}}Bounding Box \\ Dist.$\downarrow$\end{tabular}}}                                & \multicolumn{1}{c|}{MTCNN}   & 4.0305 ± 0.0212    & \underline{2.3112 ± 0.0112}     & 2.6804 ± 0.0165         & 2.8873 ± 0.0157         & \textbf{2.2255 ± 0.0116}  \\
   \multicolumn{1}{c|}{}                                                                                                                          & \multicolumn{1}{c|}{Dlib}    & 6.9492 ± 0.0692    & \underline{4.2758 ± 0.0465}     & 6.2201 ± 0.0671         & 5.7679 ± 0.0565         & \textbf{3.6719 ± 0.0413}  \\ \midrule
   \multicolumn{1}{c|}{\multirow{2}{*}{\begin{tabular}[c]{@{}c@{}}Landmark  \\  Dist.$\downarrow$\end{tabular}}}                                  & \multicolumn{1}{c|}{MTCNN}   & 9.1788 ± 1.1942   & \underline{5.8913 ± 0.0817}     & 11.4558 ± 0.4103        & 12.5190 ± 0.5016        & \textbf{3.7130 ± 0.0383}  \\
   \multicolumn{1}{c|}{}                                                                                                                          & \multicolumn{1}{c|}{Dlib}    & 10.4545 ± 1.0196   & \underline{5.5915 ± 0.0341}     & 9.2113 ± 0.4001         & 11.9333 ± 0.6150        & \textbf{3.8802 ± 0.0339}  \\ \midrule
   \multicolumn{1}{c|}{\multirow{2}{*}{\begin{tabular}[c]{@{}c@{}}Shape  \\  Dist.$\downarrow$\end{tabular}}}                                     & \multicolumn{1}{c|}{DCEA}    & 10.6392 ± 0.5002   & \underline{10.1948 ± 0.0512}    & 27.5920 ± 0.9010        & 89.4579 ± 2.0150        & \textbf{3.4060 ± 0.0294}  \\
   \multicolumn{1}{c|}{}                                                                                                                          & \multicolumn{1}{c|}{D3DFR}   & \underline{0.3429 ± 0.0014}   & 0.3591 ± 0.0061     & 0.5198 ± 0.0022         & 0.3731 ± 0.0034         & \textbf{0.3331 ± 0.0010}  \\ \midrule
   \multicolumn{1}{c|}{\multirow{2}{*}{\begin{tabular}[c]{@{}c@{}}Expression  \\  Dist.$\downarrow$\end{tabular}}}                                & \multicolumn{1}{c|}{DCEA}    & \underline{8.7972 ± 0.0235}   & 9.4759 ± 0.0474     & 27.5920 ± 1.0150        & 82.3905 ± 3.0125        & \textbf{3.0625 ± 0.0006}  \\
   \multicolumn{1}{c|}{}                                                                                                                          & \multicolumn{1}{c|}{D3DFR}   & \underline{0.1037 ± 0.0001}   & 0.1125 ± 0.0078     & 0.1757  ± 0.0006        & 0.1477 ± 0.0002         & \textbf{0.0936 ± 0.0001}  \\ \midrule
   \multicolumn{1}{c|}{\multirow{2}{*}{\begin{tabular}[c]{@{}c@{}}Pose  \\  Dist.$\downarrow$\end{tabular}}}                                      & \multicolumn{1}{c|}{HOPENet} & 2.7625  ± 0.0127   & \underline{2.1596 ± 0.0018}     & 2.3746  ± 0.0056        & 2.2970 ± 0.0061         & \textbf{2.0113 ± 0.0058}           \\
   \multicolumn{1}{c|}{}                                                                                                                          & \multicolumn{1}{c|}{FSANet}  & 38.3855 ± 1.1247   & \underline{18.0165 ± 0.9251}    & 30.3878 ± 0.9811        & 35.8777 ± 1.6530        & \textbf{13.7637 ± 0.4036} \\ \bottomrule
   \end{tabular}}
   \label{tab:utility}
   \end{table*}

\textbf{Face Utility Analysis.}
Apart from face anonymization, it is also crucial for a model to maintain the data utility of the anonymized face images.
To evaluate the utility maintenance capability of different methods, we choose different metrics for evaluating utility for different downstream tasks, including face detection, bounding box detection, landmark detection, face shape estimation, expression recognition, and pose estimation.
Concretely, we measure the utility of generated anonymized faces by using three 2D and three 3D face image metrics for a comprehensive evaluation.
For the 2D face metrics, we employ two face detection libraries, MTCNN \cite{zhang2016joint} and Dlib \cite{king2009dlib}, to evaluate the face detection rate, bounding box distance, and landmark distance between the original and anonymized faces.
In terms of the 3D face metrics, we use two 3D face reconstruction models, DCEA \cite{feng2021learning} and D3DFR \cite{deng2019accurate}, to calculate the distance of the shape embeddings and expression embeddings between the original and anonymized faces.
Additionally, we adopt two head pose estimation models, HOPENet \cite{ruiz2018fine} and FSANet \cite{yang2019fsa}, to compute the pose distance of the original and anonymized faces.
Here, we choose the L2 distance metric for all the above distance calculations.
Moreover, we also involve the image fidelity metric by comparing the Fr$\mathrm{\acute{e}}$chet Inception Distance (FID) scores~\cite{heusel2017gans} of the generated results among different methods.

Table~\ref{tab:utility} tabulates the utility comparison results of different methods.
Compared with other methods, our proposed G\textsuperscript{2}Face shows prominent advantages in terms of FID, bounding box distance, landmark distance, shape distance, expression distance, and pose distance.
For the face detection rate, our method also achieves comparable performance to state-of-the-art methods such as RiDDLE~\cite{li2023riddle} and FALCO \cite{barattin2023attribute}.
Considering all the above, our method can not only perform face anonymization precisely but also excellently maintain the utility for downstream tasks, demonstrating its huge potential in real-world applications.

\begin{table}[]
\centering
   \caption{The bit error rate and float MSE error of password extraction on CelebA-HQ.}
\begin{tabular}{@{}cc@{}}
\toprule
Bit Error Rate & Float MSE Error \\ \midrule
0.0001         & 0.0009          \\ \bottomrule
\end{tabular}
\label{tab: bit}
\end{table}

\begin{table}[]
   \centering
   \caption{Image quality comparison of recovered face images among different methods on the CelebA-HQ dataset.}
   \begin{tabular}{@{}ccccc@{}}
   \toprule
   Method                         & MAE  $\downarrow$  & LPIPS $\downarrow$           & SSIM  $\uparrow$           & PSNR $\uparrow$            \\ \midrule
   FIT~\cite{gu2020password}      & 0.0718             & 0.1674                       & \underline{0.7492}         & \underline{22.3454} \\
   RiDDLE~\cite{li2023riddle}   & \underline{0.0760} & \underline{0.1054}           & 0.7488                     & 19.0949          \\ \midrule
   \textbf{Ours}                           & \textbf{0.0509}    & \textbf{0.1017}              & \textbf{0.8165}            & \textbf{22.8203}          \\ \bottomrule
   \end{tabular}
   \label{tab:recovery}
\end{table}

\subsection{Recovery Comparison with SOTA Methods}
\label{sec:recovery}

\textbf{Qualitative Comparison.}
Before executing reversible recovery, we validate the accuracy of the extracted identity password. Only upon confirming the correctness of the extracted password can we proceed with identity recovery effectively. Our assessment involves computing the bit error rate and mean squared error (MSE) when converting binary to floating-point numbers on the CelebA-HQ~\cite{karras2018progressive} dataset, as presented in Table~\ref{tab: bit}. From the results, it's evident that our approach accurately extracts the embedded binary password and converts it into valid identity information for seamless recovery. Notably, our method does not prioritize information steganography, hence we did not delve into evaluating the performance of password steganography further.

Besides the face anonymization, we also compare our method with two representative approaches in terms of face recovery: FIT~\cite{gu2020password} and RiDDLE~\cite{li2023riddle}.
We showcase the anonymization and recovery results of different methods in Fig.~\ref{fig:visual_comparison}.
We can observe that although FIT~\cite{gu2020password} can recover the original identity, it undergoes significant facial details loss and noticeable complexion changes. RiDDLE~\cite{li2023riddle} is capable of generating vivid facial patterns but struggles to recover the original face due to its GAN inversion-based scheme.
In comparison, our method can perfectly reconstruct the original face with vivid and faithful facial details.

\begin{figure}[t]
    \centering
    \subfloat[Input]{
        \begin{minipage}{0.18\linewidth}
            \includegraphics[height=\linewidth]{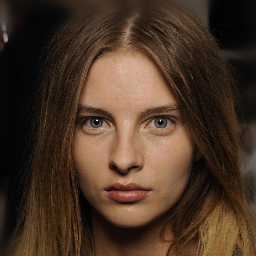} \\
            \includegraphics[height=\linewidth]{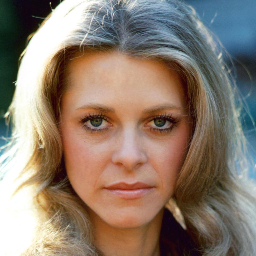} \\
            \includegraphics[height=\linewidth]{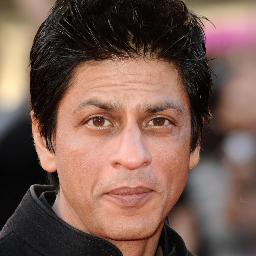} \\
        \end{minipage}
    }
    \hspace{-12pt}
    \subfloat[$Y_1$]{
        \begin{minipage}{0.18\linewidth}
            \includegraphics[height=\linewidth]{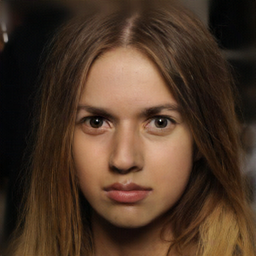} \\
            \includegraphics[height=\linewidth]{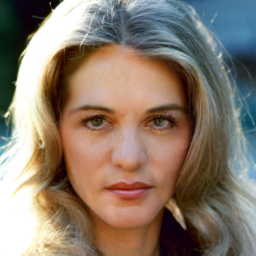} \\
            \includegraphics[height=\linewidth]{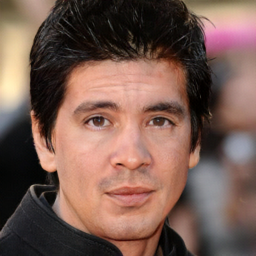} \\
        \end{minipage}
    }
    \hspace{-12pt}
    \subfloat[$Y_2$]{
        \begin{minipage}{0.18\linewidth}
            \includegraphics[height=\linewidth]{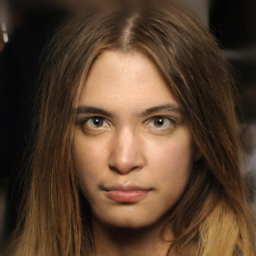} \\
            \includegraphics[height=\linewidth]{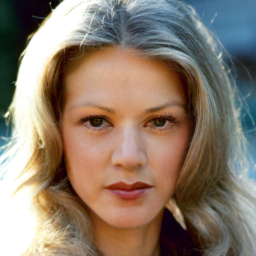} \\
            \includegraphics[height=\linewidth]{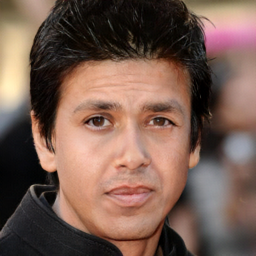} \\
        \end{minipage}
    }
    \hspace{-12pt}
    \subfloat[$Y_3$]{
        \begin{minipage}{0.18\linewidth}
            \includegraphics[height=\linewidth]{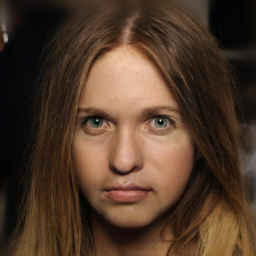} \\
            \includegraphics[height=\linewidth]{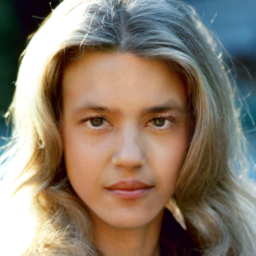} \\
            \includegraphics[height=\linewidth]{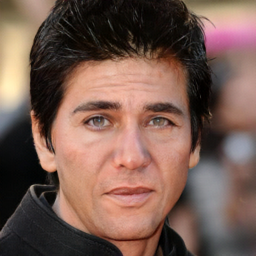} \\
        \end{minipage}
    }
    \hspace{-12pt}
    \subfloat[Zoom in]{
        \begin{minipage}{0.27\linewidth}
            \includegraphics[height=\linewidth]{figs/subfigs/multi_0_0.png} \\
            \includegraphics[height=\linewidth]{figs/subfigs/multi_0_1.png} \\
        \end{minipage}
    }
    \caption{
        Visual results of diverse anonymized counterparts for the same input face given different ID embeddings on the CelebA-HQ dataset. Zoom in for better visualization.}
     \label{fig:diversity}
\end{figure}

\textbf{Quantitative Comparison.}
To further evaluate the recovery performance of our method quantitatively, we compare our method with FIT~\cite{gu2020password} and RiDDLE~\cite{li2023riddle} on several image quality metrics, including the MAE, LPIPS~\cite{zhang2018unreasonable}, SSIM\cite{wang2004image}, and PSNR~\cite{hore2010image} between the recovered and the original input face images. The corresponding results are shown in Table~\ref{tab:recovery}. We can see that our method surpasses the compared methods in all four metrics, demonstrating its superior recovery performance. This is consistent with the qualitative comparison results in Figs.~\ref{fig:visual_comparison}((g)-(i)).
Moreover, it is also worth noting that our method elaborately embeds the password embedding into the anonymized face without compromising the anonymization/recovery quality, thereby evading the need for a troublesome pre-defined password for recovery that is required by both FIT~\cite{gu2020password} and RiDDLE~\cite{li2023riddle}. This feature grants our method better scalability and makes it more practical and convenient for real-world applications.

\subsection{Anonymization Diversity Analysis}
\label{sec:diversity}

   Generation diversity is a key indicator to measure the performance of face anonymization approaches. To evaluate the diversity of the generated anonymized faces of our method, we showcase the anonymization results of input faces given different dummy IDs in Fig.~\ref{fig:diversity}.
   As can be seen, our method can effectively deliver anonymized faces with diverse identities and realistic facial details, demonstrating the compelling diverse generation capability of our method.

\begin{figure}[t]
    \centering
    \subfloat[CIAGAN~\cite{maximov2020ciagan}]{
        \vspace{-20pt}
        \begin{minipage}{0.4\linewidth}
            \includegraphics[height=\linewidth]{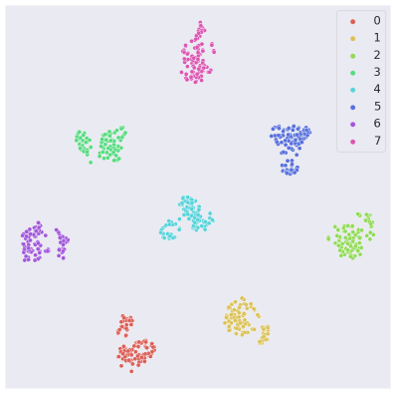} \\
        \end{minipage}
    }
    \subfloat[FIT~\cite{gu2020password}]{
        \vspace{-20pt}
        \begin{minipage}{0.4\linewidth}
            \includegraphics[height=\linewidth]{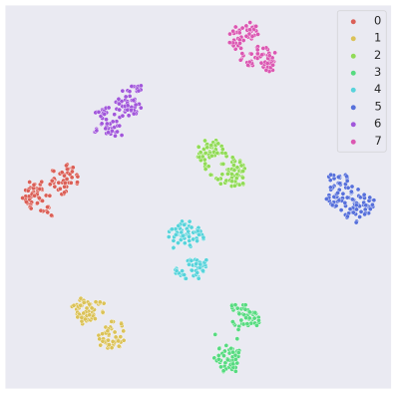} \\
        \end{minipage}
    }
    \vspace{-6pt}
    
    \subfloat[RiDDLE~\cite{li2023riddle}]{
        \vspace{-12pt}
        \begin{minipage}{0.4\linewidth}
            \includegraphics[height=\linewidth]{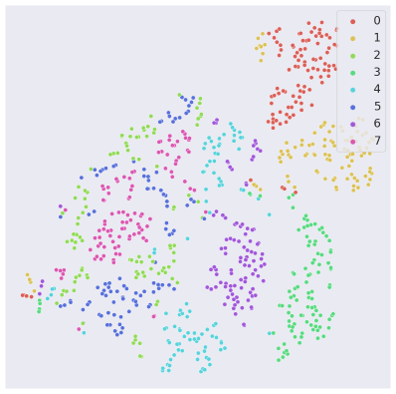} \\
        \end{minipage}
    }
    \subfloat[Ours]{
        \vspace{-12pt}
        \begin{minipage}{0.4\linewidth}
            \includegraphics[height=\linewidth]{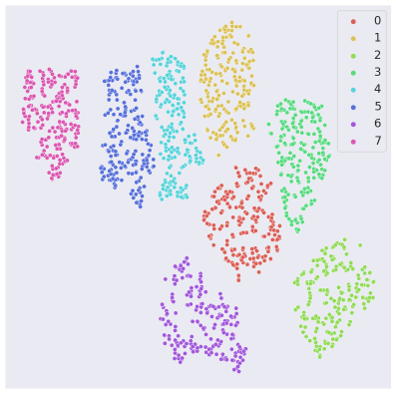} \\
        \end{minipage}
    }
    \caption{
        t-SNE visualization of identity embeddings of anonymized images generated by different methods.}
      \label{fig:tsne}
\end{figure}

   For a more thorough comparison, we also compare the generation diversity of our method with the existing methods by visualizing the identity embeddings of anonymized faces. Initially, we randomly select 8 face images from the testing data of the CelebA-HQ dataset. Then, we generate 200 anonymized faces for each selected face image by sampling random dummy IDs.
   Afterward, we extract the identity embeddings of the anonymized images using a pre-trained ArcFace\cite{deng2019arcface} model and perform t-SNE visualization of these embeddings~\cite{van2008visualizing} in Fig.~\ref{fig:tsne}.
   As is observed, the identities generated by CIAGAN~\cite{maximov2020ciagan} and FIT~\cite{gu2020password} distribute in a relatively small area for each cluster, indicating that the anonymized images generated by these two methods exhibit low diversity.
   In comparison, RiDDLE~\cite{li2023riddle} generates more distributed identity embeddings. The rationale behind this is that RiDDLE~\cite{li2023riddle} relies on the GAN inversion to generate anonymized images, which tends to manipulate the identity information with significant changes on the other id-irrelevant attributes of the input face. This results in a higher diversity of the generated anonymized images at the cost of severe deterioration in the utility of the anonymized images.
   Unlike all the above methods, our approach not only generates more distributed identities but also tends to cluster in proximity when the embeddings are derived from the same source. This indicates that our method can generate diverse anonymized faces while simultaneously preserving the geometric consistency with the original face, demonstrating its superior performance in terms of both face anonymization and utility maintenance.

\begin{figure*}[t]
    \centering
    \captionsetup[subfloat]{font=tiny}
    \subfloat[Input]{
        \begin{minipage}{0.1\linewidth}
            \includegraphics[height=\linewidth]{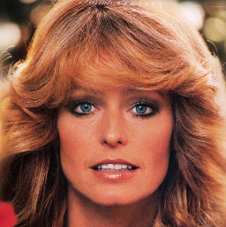} \\
            \includegraphics[height=\linewidth]{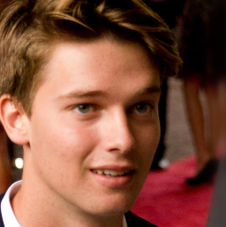} \\
        \end{minipage}
    }
    \hspace{-12pt}
    \subfloat[GENP]{
        \begin{minipage}{0.1\linewidth}
            \includegraphics[height=\linewidth]{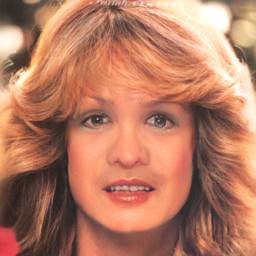} \\
            \includegraphics[height=\linewidth]{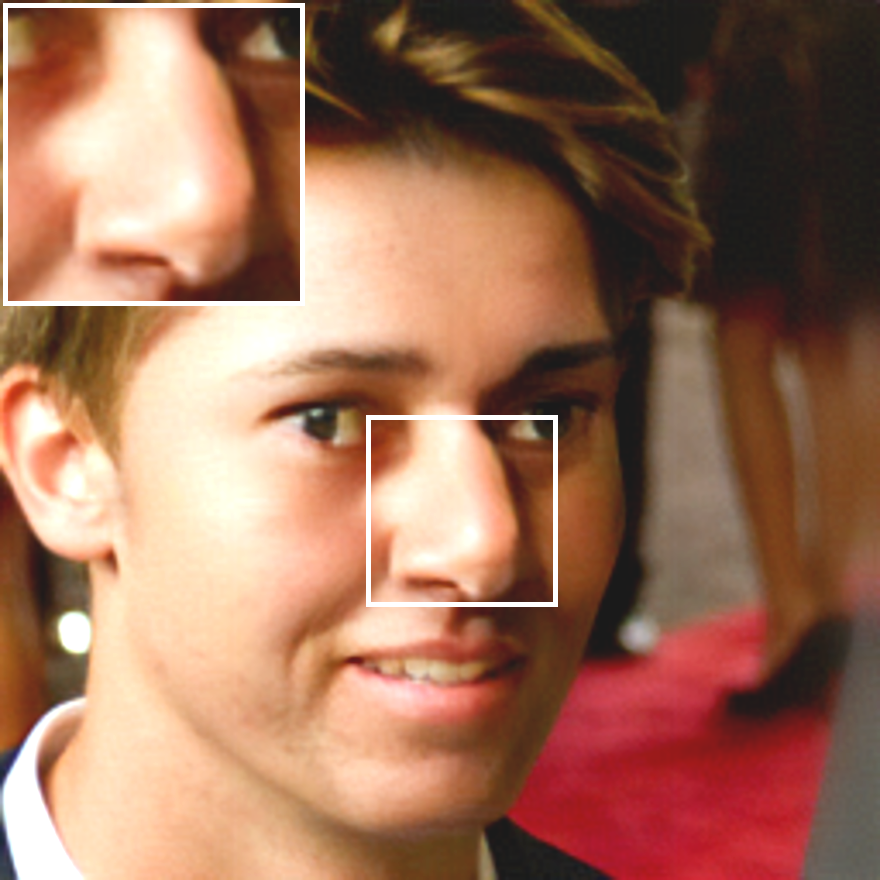} \\
        \end{minipage}
    }
    \hspace{-12pt}
    \subfloat[GEOP]{
        \begin{minipage}{0.1\linewidth}
            \includegraphics[height=\linewidth]{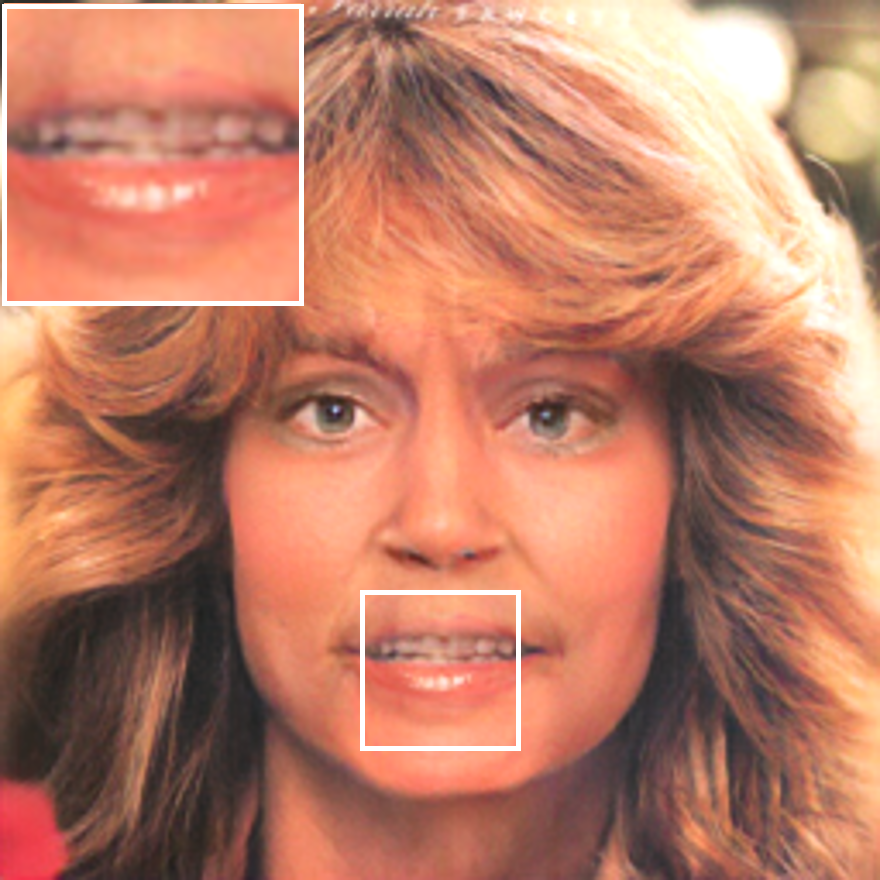} \\
            \includegraphics[height=\linewidth]{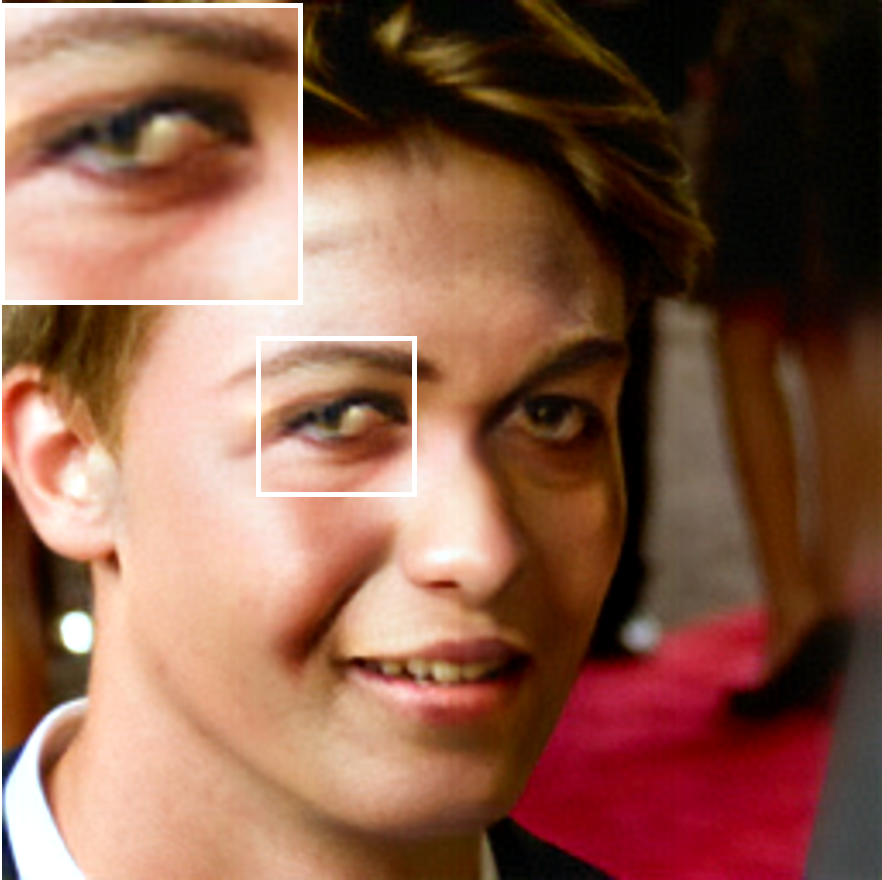} \\
        \end{minipage}
    }
    \hspace{-12pt}
    \subfloat[IFF]{
        \begin{minipage}{0.1\linewidth}
            \includegraphics[height=\linewidth]{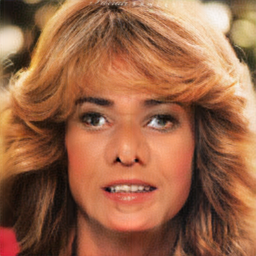} \\
            \includegraphics[height=\linewidth]{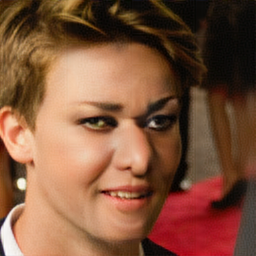} \\
        \end{minipage}
    }
    \hspace{-12pt}
    \subfloat[GEOP + IFF]{
        \begin{minipage}{0.1\linewidth}
            \includegraphics[height=\linewidth]{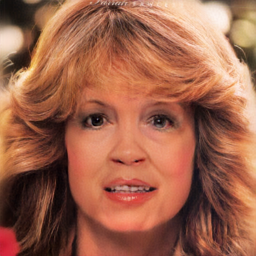} \\
            \includegraphics[height=\linewidth]{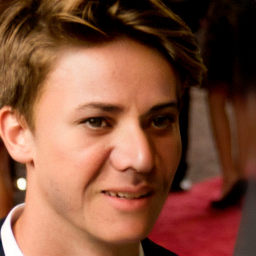} \\
        \end{minipage}
    }
    \hspace{-12pt}
    \subfloat[GENP + IFF]{
        \begin{minipage}{0.1\linewidth}
            \includegraphics[height=\linewidth]{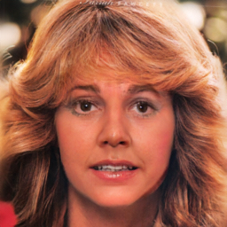} \\
            \includegraphics[height=\linewidth]{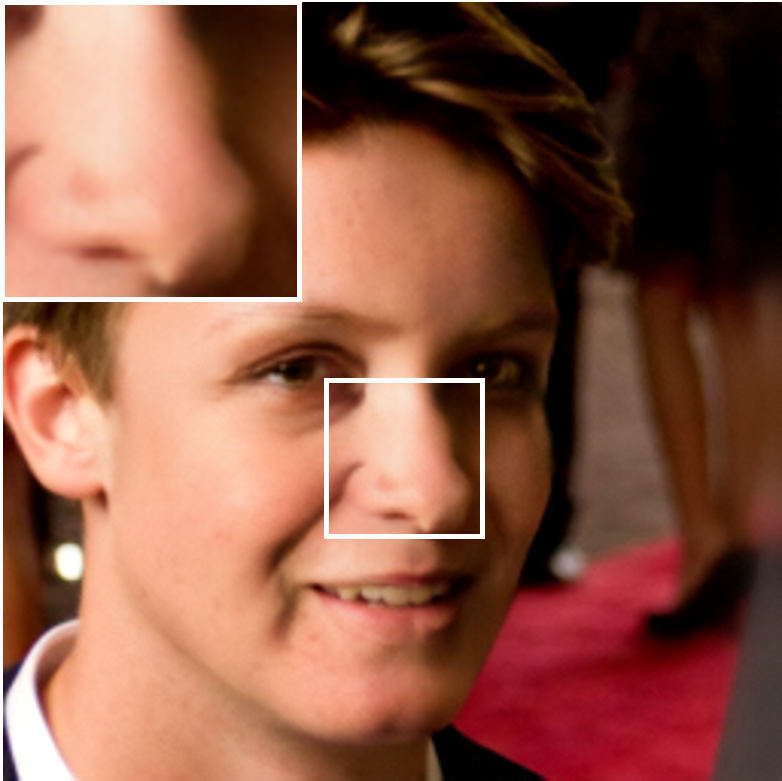} \\
        \end{minipage}
    }
    \hspace{-12pt}
    \subfloat[GENP + GEOP]{
        \begin{minipage}{0.1\linewidth}
            \includegraphics[height=\linewidth]{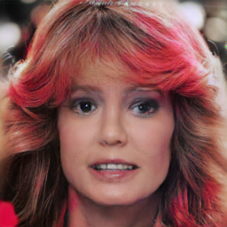} \\
            \includegraphics[height=\linewidth]{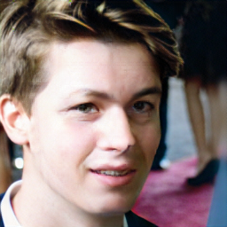} \\
        \end{minipage}
    }
    \hspace{-12pt}
    \subfloat[Full(Ours)]{
        \begin{minipage}{0.1\linewidth}
            \includegraphics[height=\linewidth]{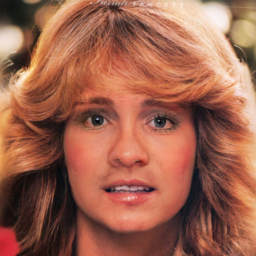} \\
            \includegraphics[height=\linewidth]{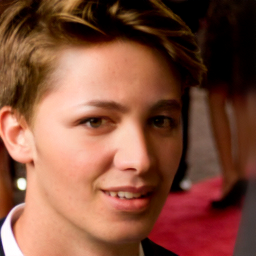} \\
        \end{minipage}
    }
    \caption{
        Qualitative ablation study on the CelebA-HQ dataset.}
     \label{fig:ablation}
\end{figure*}

\subsection{Ablation Study}
\label{sec:ablation}

\begin{table}[t]
\centering
\caption{Quantitative ablation study on the CelebA-HQ dataset.}
\setlength{\tabcolsep}{3pt}
\begin{tabular}{@{}ccc|cccccc@{}}
\toprule
GENP       & GEOP       & IFF        & Ano.$\downarrow$ & Rec.$\uparrow$  & FID$\downarrow$ & LM.$\downarrow$ & Shape$\downarrow$ & Exp.$\downarrow$ \\ \midrule
\checkmark &            &            & 0.1267           & 0.5339          & 25.2675         & 5.2303          & 0.4291            & 0.1140           \\
           & \checkmark &            & 0.1396           & 0.4705          & 42.1927         & 4.9960          & 0.4304            & 0.1283           \\
           &            & \checkmark & 0.1411           & 0.4187          & 54.2915         & 7.9324          & 0.4401            & 0.1379           \\
           & \checkmark & \checkmark & 0.1234           & 0.5095          & 36.8143         & 4.1941          & 0.4011            & 0.1123           \\
\checkmark &            & \checkmark & \textbf{0.0957}  & 0.5593          & 16.2711         & 4.8523          & 0.4118            & 0.1203           \\
\checkmark & \checkmark &            & 0.1112           & {\ul 0.6912}    & {\ul 14.0371}   & {\ul 4.1917}    & {\ul 0.3479}      & {\ul 0.1034}     \\
\checkmark & \checkmark & \checkmark & {\ul 0.1055}     & \textbf{0.7954} & \textbf{7.0412} & \textbf{3.8802} & \textbf{0.3331}   & \textbf{0.0936}  \\ \bottomrule
\end{tabular}
\label{tab:ablation}
\end{table}

In this section, we present an in-depth analysis of the effectiveness of our main proposals. Particularly, we investigate the performance of different model variants by all combinations of different components, respectively. Here, we include six metrics for quantitative evaluation: the cosine similarities between the identity embeddings of the anonymized face and the original face (Ano.), and between the identity embeddings of the recovered face and the original face (Rec.) are calculated for measuring face anonymization/recovery performance. Additionally, the Fréchet Inception Distance (FID)\cite{heusel2017gans}, distances of landmarks (LM.), shape (Shape), and expressions (Exp.) are also involved in evaluating image quality and data utility maintenance, respectively.

\textbf{GENP (Generative Prior).} 
The effectiveness of GENP is verified in Fig.~\ref{fig:ablation}(b). While incorporating GENP can yield images with higher fidelity, the absence of geometric prior and the IFF block result in a noticeable gap in geometric consistency and color compared to the original image. Table \uppercase\expandafter{\romannumeral6} further confirms that GENP alone fails to produce satisfactory anonymization and reversible recovery results. 

\textbf{GEOP (Geometric Prior).} 
The effectiveness of using only GEOP is examined in Fig.~\ref{fig:ablation}(c). This variant generates images with better geometric consistency. However, the lack of GENP and IFF results in noticeable artifacts and color differences. Table~\ref{tab:ablation} also indicates higher FID scores for images generated by GEOP alone, indicating poorer image quality. 

\textbf{IFF.}
Fig.~\ref{fig:ablation}(d) presents the results of using only IFF by removing the input geometric information and pretrained weights of StyleGAN2. Although the anonymized image is more consistent in color with the original image, the outcomes exhibit obscure facial textures and notable distortions. The inherent limitations of a simple encoder-decoder network hinder high-fidelity identity manipulation, leading to inferior anonymization/recovery quality and data utility, as shown in Table~\ref{tab:ablation}. 

\textbf{GEOP + IFF.}
The effectiveness of using GEOP + IFF is shown in Fig.~\ref{fig:ablation}(e). This combination can generate geometrically consistent results but also results in overly smooth facial patterns and higher FID scores due to the lack of generative prior. This deficiency is also reflected in Table~\ref{tab:ablation}, indicating a decline in face recovery quality. 

\textbf{GENP + IFF.}
Fig.~\ref{fig:ablation}(f) shows the results of the model variant ``GENP + IFF'', which reveal that the combination of GENP and IFF can generate high-fidelity results. However, there is a notable degradation in geometric consistency (\eg, face shape or nose structure) due to the absence of the geometric prior. Table~\ref{tab:ablation} underscores the decline in utility and performance metrics when omitting the geometric prior, further evidenced by structural distortions in generated faces. 

\textbf{GENP + GEOP.}
The results of variant ``GENP + GEOP'' without the IFF block in Fig.~\ref{fig:ablation}(g) highlights the crucial role of IFF in preserving ID-irrelevant features. Drastic changes in hair and background color emphasize IFF's significance in enabling precise face anonymization. 

\textbf{Full Model.}
Fig.~\ref{fig:ablation}(h) and Table~\ref{tab:ablation} show the results when we effectively fuse all the key components proposed in our paper. Our full model produces the best results in terms of image quality, geometric consistency, and reversible recovery, indicating the effectiveness of our proposed method.

\subsection{Complexity Analysis}
We compare the computational complexity of different methods on an image with resolution $256 \times 256$ in Table~\ref{tab:complexity}. Here, three metrics, including FLOPs, Parameter size, and inference time are involved. Similar to  StyleGAN-based methods~\cite{li2023riddle, barattin2023attribute}, our computational costs (FLOPs) and model size (Params.) are higher than those of encoder-decoder networks~\cite{maximov2020ciagan, gu2020password}. Nevertheless, the inference time (Time) of our method is comparable to other approaches, enabling real-time applications. Considering our remarkable advantages in face anonymization and recovery performance, we believe such overhead is acceptable.

\begin{table}[]
\setlength{\tabcolsep}{3pt}
\caption{Computational complexity comparison of different methods on an image with resolution $256 \times 256$.}
\centering
\begin{tabular}{@{}cccccc@{}}
\toprule
                        & CIAGAN~\cite{maximov2020ciagan}         & FIT~\cite{gu2020password}           & RiDDLE~\cite{li2023riddle}  & FALCO~\cite{barattin2023attribute}   & Ours    \\ \midrule
FLOPs (G) $\downarrow$   & \textbf{38.300}  & {\ul 120.381} & 235.841 & 150.898 & 267.471 \\
Params. (M) $\downarrow$ & {\ul 11.801} & \textbf{11.433} & 40.029  & 30.370  & 81.185  \\
Time (s) $\downarrow$    & \textbf{0.022}  & {\ul 0.025}   & 0.027   & 7.661   & 0.034   \\ \bottomrule
\end{tabular}
\label{tab:complexity}
\end{table}

\subsection{Limitations and Discussions}
\label{sec:limitation}
This paper introduces G\textsuperscript{2}Face, a reversible face anonymization method that leverages both generative and geometric priors. By preserving the fidelity of generated faces and ensuring geometric consistency in anonymized faces, G\textsuperscript{2}Face enhances the utility of anonymized data. Our study includes comprehensive experiments, spanning visualization, quantification of anonymization and recovery, diversity analysis of anonymized faces, and ablation study, all demonstrating the efficacy of our approach.

\begin{figure}[t]
    \centering
    \subfloat[]{
        \begin{minipage}{0.2\linewidth}
            \includegraphics[height=\linewidth]{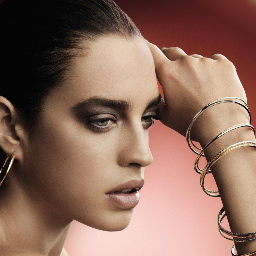} \\
            \includegraphics[height=\linewidth]{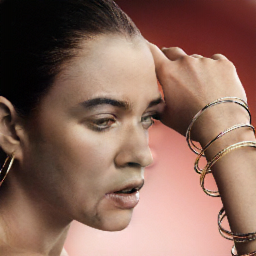} \\
        \end{minipage}
    }
    \hspace{-10.5pt}
    \subfloat[]{
        \begin{minipage}{0.2\linewidth}
            \includegraphics[height=\linewidth]{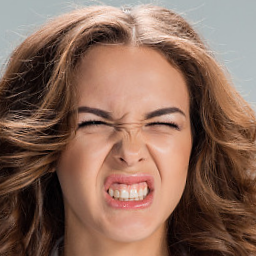} \\
            \includegraphics[height=\linewidth]{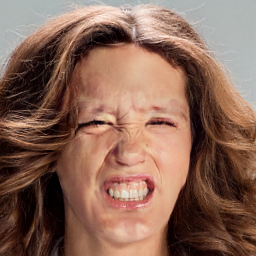} \\
        \end{minipage}
    }
    \hspace{-10.5pt}
    \subfloat[]{
        \begin{minipage}{0.2\linewidth}
            \includegraphics[height=\linewidth]{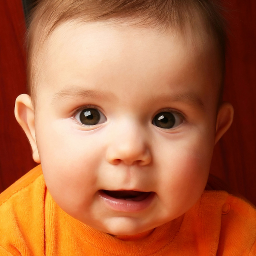} \\
            \includegraphics[height=\linewidth]{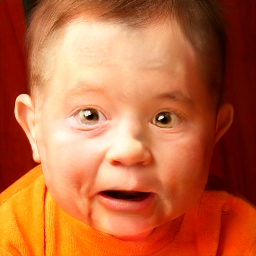} \\
        \end{minipage}
    }
    \hspace{-10.5pt}
    \subfloat[]{
        \begin{minipage}{0.2\linewidth}
            \includegraphics[height=\linewidth]{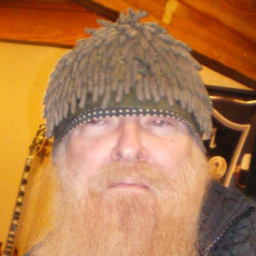} \\
            \includegraphics[height=\linewidth]{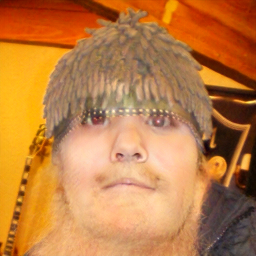} \\
        \end{minipage}
    }
    \caption{
        Failure cases of our model.}
    \label{fig:limitation}
\end{figure}

While our method exhibits superior performance in face anonymization and reversible recovery, it may yield inferior results (\eg, notable facial distortions, artifacts, and attribute loss) when encountering faces with extreme poses, expressions, aging patterns, and face attributes, as depicted in Fig.~\ref{fig:limitation}. This is primarily due to the scarcity of samples representing such scenarios in our training data. However, these limitations can be addressed by incorporating a more diverse range of data for training.

\section{Conclusion}
In this paper, we present G\textsuperscript{2}Face, a novel framework for achieving high-fidelity and geometrically consistent reversible face anonymization. 
The proposed G\textsuperscript{2}Face framework explicitly extracts the 3D information as geometric prior, which is then combined with the generative priors in a pre-trained StyleGAN in a multi-scale manner via a set of elaborately  designed identity-aware feature fusion blocks. Extensive qualitative and quantitative experiments demonstrate that the proposed method surpasses the state-of-the-art competitions in reversible face recovery with high data utility.

\bibliographystyle{IEEEtran}
\bibliography{refs}

\end{document}